\documentclass[preprint,12pt,authoryear]{elsarticle}

\usepackage{amssymb}
\newcommand{\eg}{{\em e.g.}}           
\newcommand{\ie}{{\em i.e.}}           

\usepackage{mathrsfs}
\usepackage[normalem]{ulem}
\usepackage[symbol]{footmisc}
\usepackage{subfig}
\usepackage{bbm}

\usepackage{booktabs}
\usepackage{algorithm,algorithmic}
\usepackage{multirow}
\usepackage{rotating}
\usepackage{amsmath,amsfonts,bm}
\usepackage{hyperref}

\journal{Medical Image Analysis}

\usepackage{xcolor}

\begin{document}

\begin{frontmatter}



\title{Transferring Ultrahigh-Field Representations for Intensity-Guided Brain Segmentation of Low-Field Magnetic Resonance Imaging}



\author[inst1]{Kwanseok Oh\textsuperscript{$*$,}}
\author[inst1]{Jieun Lee\textsuperscript{$*$,}}
\author[inst1]{Da-Woon Heo}
\author[inst3,inst4]{Dinggang Shen}
\author[inst1,inst2]{Heung-Il Suk\textsuperscript{$\dagger$,}}

\affiliation[inst1]{organization={Department of Artificial Intelligence, Korea University},addressline={Seoul 02841, Republic of Korea}}
\affiliation[inst2]{organization={Department of Brain and Cognitive Engineering, Korea University},addressline={Seoul 02841, Republic of Korea}}
\affiliation[inst3]{organization={School of Biomedical Engineering, ShanghaiTech University},addressline={Shanghai 202150, China},}
\affiliation[inst4]{organization={Shanghai United Imaging Intelligence Co., Ltd.},addressline={Shanghai 202150, China}}

\begin{abstract}
\footnotetext[1]{Both authors equally contributed to this research.}
\footnotetext[2]{Corresponding author: Heung-Il Suk (hisuk@korea.ac.kr)}
Ultrahigh-field (UHF) magnetic resonance imaging (MRI), \ie, 7T MRI, provides superior anatomical details of internal brain structures owing to its enhanced signal-to-noise ratio and susceptibility-induced contrast. However, the widespread use of 7T MRI is limited by its high cost and lower accessibility compared to low-field (LF) MRI. This study proposes a deep-learning framework that systematically fuses the input LF magnetic resonance feature representations with the inferred 7T-like feature representations for brain image segmentation tasks in a 7T-absent environment. Specifically, our adaptive fusion module aggregates 7T-like features derived from the LF image by a pre-trained network and then refines them to be effectively assimilable UHF guidance into LF image features. Using intensity-guided features obtained from such aggregation and assimilation, segmentation models can recognize subtle structural representations that are usually difficult to recognize when relying only on LF features. Beyond such advantages, this strategy can seamlessly be utilized by modulating the contrast of LF features in alignment with UHF guidance, even when employing arbitrary segmentation models. Exhaustive experiments demonstrated that the proposed method significantly outperformed all baseline models on both brain tissue and whole-brain segmentation tasks; further, it exhibited remarkable adaptability and scalability by successfully integrating diverse segmentation models and tasks. These improvements were not only quantifiable but also visible in the superlative visual quality of segmentation masks.
\end{abstract}


\end{frontmatter}

\section{Introduction}\label{section:intro}
Brain image segmentation is a pivotal process in medical imaging that offers a foundation for advanced volumetric analysis, diagnosis, and treatment planning. Magnetic resonance imaging (MRI), which is considered one of the most crucial noninvasive scanning techniques, provides detailed insights into internal brain structures \citep{clarke1995mri}. The accurate delineation of brain structures from MRI scans allows clinicians and researchers to identify pathological areas, enabling the precise localization of lesions, tumors, and other abnormalities~\citep{icsin2016review,jyothi2023deep}. In this regard, the series of brain segmentation is roughly divided into two primary studies: brain tissue \citep{chen2018voxresnet, wang2019rp, li2023can} and whole-brain \citep{roy2019quicknat,henschel2020fastsurfer,yu2023unest} segmentation. Brain tissue segmentation involves partitioning a brain image into its constituent tissue types, typically categorized as gray matter (GM), white matter (WM), and cerebrospinal fluid. By contrast, whole-brain segmentation extends this approach to include all discernible brain structures to generate a comprehensively parcellated map of the brain's anatomy.
Both segmentation tasks have been intensively studied using low-field (LF) magnetic resonance (MR) images, such as 1.5T or 3T MRI, owing to their prevalence and cost-effectiveness in the clinic. However, 1.5T/3T images present a challenge due to the lower signal-to-noise ratio (SNR) and spatial contrast resolution compared to ultrahigh-field (UHF) MRI scanners. UHF MR images such as 7T MRI offer enhanced visualization of nuanced anatomical details~\citep{shaffer2022ultra} with their higher SNR and image resolution~\citep{heiss2023clinical}, which is particularly beneficial for identifying subtle brain pathologies regarding abnormalities. In addition to these favorable properties, with a strong tissue contrast, 7T images tend to detect even morphological variations that are arduous to perceive in 3T images, \ie, significant differences in voxel intensity. Thus, the utilization of 7T images not only allows advanced brain-tissue segmentation but could also be a breakthrough in enhancing medical image analysis performance within the whole brain, including cortical and subcortical areas~\citep{bazin2014computational,svanera2021cerebrum,miletic2022charting}. However, 7T MRI acquisition is significantly more expensive, making its usage in practical clinical environments difficult. 

\begin{figure}[t!]
\centering
\includegraphics[width=\textwidth]{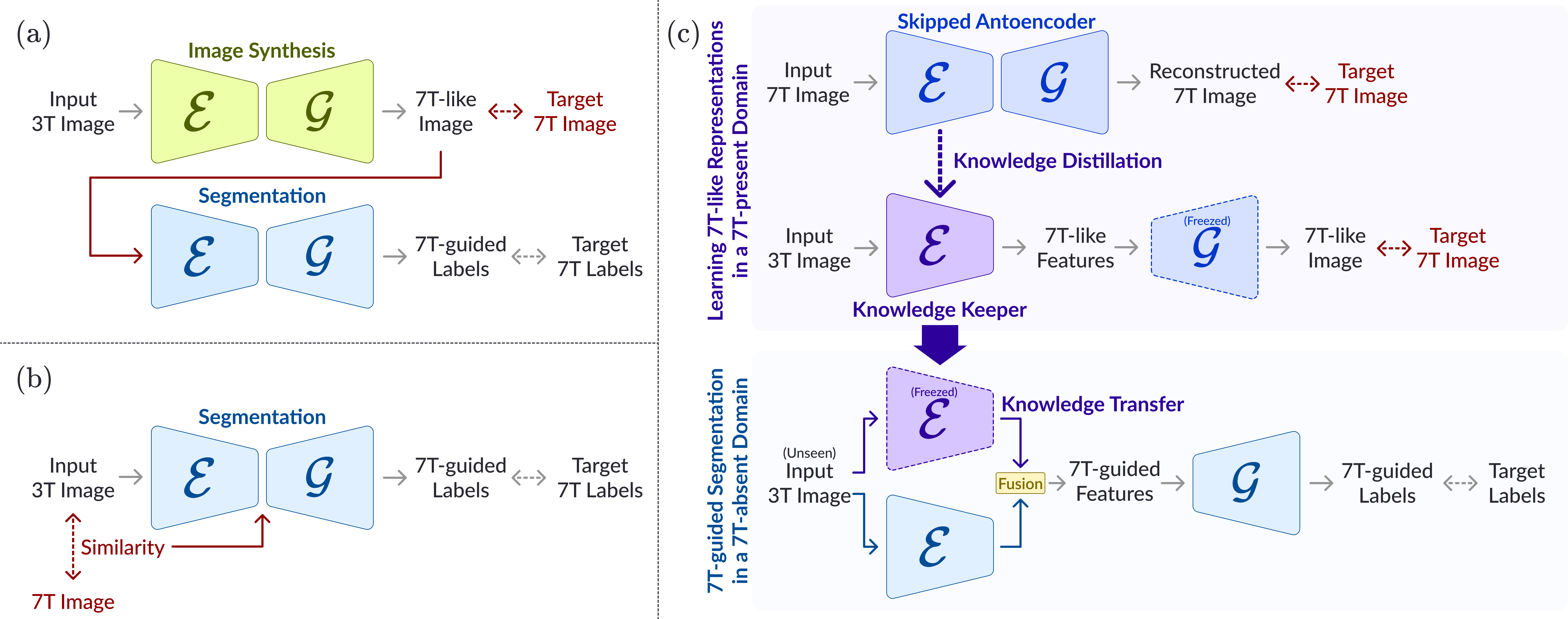}
\caption{Conceptual differences between conventional learning-based 3T-to-7T methods using (a) a 7T-like image as the guidance, (b) directly matched features as the guidance, and (c) the proposed method. Here, $\mathcal{E}$ and $\mathcal{G}$ denote an encoder and decoder, respectively.}\label{concept_feature_transfer}
\end{figure}

Meanwhile, deep learning (DL) techniques have achieved unprecedented success in the medical field, with their remarkable capability to learn complex structural patterns and feature representations from medical imaging data. With the incredible power of DL, there have been efforts to take advantage of UHF MRIs, despite their limited availability, by designing 7T-guided brain segmentation models, particularly by translating 3T images into UHF representations (\ie, 7T-like feature representations). Specifically, there are two different approaches in the existing 3T-to-7T translation methods: (i) using a 7T-like image as the guidance (Figure~\ref{concept_feature_transfer}(a)) and (ii) using directly matched features as the guidance (Figure~\ref{concept_feature_transfer}(b)). The first approach \citep{zhang2018dual,qu2020synthesized,duan2023synthesized} involves transforming the appearance of 3T images to resemble that of the corresponding 7T images (\ie, 7T-like image translation). This approach leverages the superior contrast and detail of 7T images to enhance the quality of 3T images; however, it presents the difficulty of accurately mimicking complex 7T characteristics, and the inherent 3T features are prone to unexpected deformation or distortion. The second approach \citep{deng2016learning,wei2021cascaded} focuses on extracting hierarchical features by directly matching 3T and 7T images. By utilizing handcrafted or task-dependent features derived from the similarity of paired images, this approach benefits from using abundant information unique in 7T characteristics. However, these features might be limited in the robustness and diversity of UHF representations owing to biasing toward the pre-defined features. The dependency on matched pairs of 3T and 7T images significantly compromises the efficacy of segmentation model training in both the abovementioned approaches. Therefore, such drawbacks pose a considerable barrier to facilitating feasible UHF guidance.

To overcome the aforementioned issues, in our previous work \citep{lee2022novel} we devised a knowledge keeper network (KKN) to estimate hierarchical 7T-like features using a 3T image via knowledge distillation (KD) \citep{bucilua2006model,hinton2015distilling} and provide UHF feature representations to an arbitrary segmentation model as a way of knowledge transfer (Figure~\ref{concept_feature_transfer}(c)). This strategy encouraged an independent segmentation model to harness UHF guidance by employing a weighted sum between LF and 7T-like features in a domain devoid of the 7T MRI. Although such a simple feature blending relying on two learnable parameters showed performance improvement in brain tissue segmentation, the method struggled to fully utilize UHF representations owing to the inevitable conflict of coupling between the excessive contrast of 7T-like features and inherent structures of LF images.

In the present work, we propose an adaptive fusion strategy that transfers optimal guided features for segmentation by adjusting the intensity values of LF features according to UHF guidance. To adequately fuse strong contrast information with structural LF image features, we devise an adaptive fusion module (AFM) for a knowledge flow that aggregates and assimilates UHF feature representations, \ie, new knowledge of the 7T-absent domain, as depicted in Figure~\ref{fusion_module}(a). In this way, the AFM produces UHF guidance that has positive and negative values to emphasize a particular region from LF features by making voxel intensities stronger or weaker. To this end, comprehensive guidance is first obtained from 7T-like features delivered by the pre-trained KKN by devising simple knowledge aggregation (Figure~\ref{fusion_module}(b)). The comprehensive guidance portrays holistic representations that incorporate fine details and broader contextual information. Such guidance is then altered to be individualized for the given LF features by feature-combined refinement (Figure~\ref{fusion_module}(c)). Subsequently, we separate the guidance in two ways according to the sign of the values, \ie, positive and negative, and assign learnable weights with the channel attention mechanism to produce assimilable guidance. This channel-wise refinement enables the AFM to specialize UHF guidance via the directionality and magnitude in which LF features are to be adjusted. We finally incorporate the assimilable guidance into LF features so that intensity-guided features enable the identification of specific regions as either brighter or darker than their surroundings. This process helps to enhance the contrast and detail of principal anatomical areas in LF images, bridging the gap between LF and UHF imaging qualities. 

To the best of our knowledge, this work is the first to effectively integrate intensity-adjusted UHF guidance into LF features for enhanced brain-tissue segmentation, leveraging an innovative adaptive-fusion approach. By promoting the representation of the pre-trained KKN with AFM, the proposed method allows the segmentation model to reveal imperceivable pathological variations when relying solely on LF image features. Experimental results showed that the proposed method outperformed several baselines on brain tissue and whole-brain segmentation tasks. By further investigating the quality of guided features derived from the AFM, we demonstrated that such guided features have a more sophisticated depiction with enhanced intensities while preserving the overall appearance of LF features. As such, our method adequately balances the strong contrast of UHF and structural information of LF at the feature level, achieving optimal intensity-guided segmentation.

The main contributions of this study are as follows:
\begin{itemize}
    \item We propose an AFM that assimilates UHF guidance in arbitrary segmentation models of a 7T-absent domain by adjusting the contrasts of LF image features.
    \item We introduce a knowledge flow that aggregates UHF feature representations transferred from a pre-trained network seamlessly and then refines UHF guidance to be specialized in LF features without deforming the inherent LF characteristics.
    \item We demonstrate that the proposed method leads to performance improvement quantitatively and qualitatively on two benchmark datasets for different segmentation tasks: the Internet Brain Segmentation Repository (IBSR) for brain tissue segmentation and the Multi-Atlas Labeling Challenge (MALC) for whole-brain segmentation.
\end{itemize}

\section{Related Work}
\label{section:related}

\subsection{3T-to-7T Translation Approaches}
With the great success of deep neural networks in medical imaging, learning-based methods have demonstrated the effectiveness of UHF guidance on paired 3T and 7T images by synthesizing 7T-like images \citep{bahrami20177t,zhang2018dual,qu2020synthesized,duan2023synthesized} or extracting directly matched features \citep{deng2016learning,wei2021cascaded}.

For 7T-like image synthesis, \citet{bahrami20177t} and \citet{zhang2018dual} used regression techniques to increase the correlation between 3T and 7T images. \citet{bahrami20177t} compared tissue-segmented results derived from 3T and 7T-like images by employing FMRIB's automated segmentation tool (FAST) in the FMRIB software library (FSL) package \citep{zhang2001segmentation} to demonstrate that 7T-like images afford more accurate parcellated tissue labels compared to 3T images. Meanwhile, \citet{qu2020synthesized} proposed a wavelet-based affine transformation network (WATNet) that leveraged complementary information of both spatial and wavelet domains to generate 7T-like images. They also employed a widely used segmentation pipeline—advanced normalization tools (ANTs) \citep{avants2011open}—to evaluate the influence of 7T-like images on tissue segmentation compared to the use of 3T images. However, the synthesized image is not always guaranteed to be helpful for structural segmentation because inherent features such as the appearance of the 3T image might vanish owing to the image-to-image translation with excessive contrast.

To leverage complementary information between 3T and 7T at the feature level, \citet{deng2016learning} computed a large number of three-dimensional (3D) Haar-like features for each voxel and then utilized them to refine the tissue probability maps progressively by directly matching 3T to 7T features. \citet{wei2021cascaded} developed a CaNes-Net with a correlation coefficient map, which was defined to quantitatively measure the degree of alignment between 3T and 7T images using the Pearson correlation. They focused on alleviating the misalignment of the affine transformation between two images acquired for the same subject by incorporating the map into the cross-entropy loss. In such an approach using directly matched features, the model is trained to segment the 3T image by minimizing the feature distance to the ground truth obtained from the corresponding 7T image. However, such rule-based and handcrafted feature-extraction methods, being heavily reliant on paired 3T and 7T training datasets, often lack robustness. Similarly, the approach for 7T-like image synthesis has also resulted in identical limitations, as they demand the corresponding 7T images to achieve 7T-guided results from the 3T image. To utilize UHF guidance without 7T images, the proposed method learns how to transform a 3T image into 7T-like feature representations on a paired dataset and then provides pre-trained UHF feature representations with the arbitrary segmentation model using only the LF MR image in a 7T-absent domain.

\subsection{Brain Segmentation in Structural MRI}
Brain image segmentation using structural MRI is essential in analyzing brain structures by visualizing complex and diverse anatomical regions. In neuroimaging research, conventional automated tools have long been adopted for brain image segmentation \citep{fischl2002whole, avants2011open}. However, the convolutional neural network (CNN)-based method has recently become a powerful segmentation technique with remarkable performance owing to the notable advancement of U-Net \citep{ronneberger2015u, cciccek20163d}. The distinctive encoder-decoder structure of U-Net, characterized by its U-shape and skip connections, has inspired numerous segmentation approaches to segment an image into specific tissue labels such as WM, GM, and cerebrospinal fluid. In this regard, \citet{kumar2018u} proposed U-SegNet, which combines the strengths of U-Net and SegNet \citep{badrinarayanan2017segnet} into a hybrid architecture to optimize brain tissue segmentation with a more efficient use of parameters. For the volumetric segmentation of the MR image, \citet{chen2018voxresnet} introduced a voxel-wise residual network (VoxResNet) that extended two-dimensional (2D) residual learning into a 3D network while RP-Net \citep{wang2019rp} modified the 3D U-Net by applying recursive residual blocks and a pyramid pooling module. 

Most brain segmentation tasks are confined to a few segment predictions, while whole-brain segmentation aims to divide the entire brain into a multitude of regions, including cerebral WM, GM, and various subcortical structures. Because of the intricacies involved with numerous segments, many whole-brain segmentation methods adopt a 2D slice-based approach with view aggregation \citep{roy2019quicknat,henschel2020fastsurfer} or a patch-wise approach with 3D hierarchical block aggregation \citep{yu2023unest}. Specifically, three predicted probability maps of segmentation are produced by three 2D CNNs trained respectively on each plane, namely, coronal, axial, or sagittal. These three predictions are combined in the final view aggregation stage to generate a segmentation mask, allowing the recapture of spatial information in 3D. In this way, \citet{roy2019quicknat} proposed the quick segmentation of neuroanatomy (QuickNAT) that had two consecutive steps: pre-training on large unlabeled datasets with auxiliary labels and fine-tuning on a manually annotated dataset with 27 cortical and subcortical structures. With the basic architecture of QuickNAT as the inspiration, FastSurfer \citep{henschel2020fastsurfer} was proposed with three 2D CNNs that had competitive dense encoder and decoder blocks by replacing concatenation with max-out operations. Recently, \citet{yu2023unest} have proposed a UNesT comprising a simplified and more rapidly converging transformer encoder design for 3D whole-brain segmentation.

From the abovementioned perspectives, we devise respective 3D and 2D training strategies to validate the adaptability and scalability of the proposed method on both brain tissue and whole-brain segmentation. By enforcing the proposed method on two different tasks, we demonstrate the expansive applicability of UHF guidance, as our method enables the successful transfer of UHF feature representations regardless of the 3D or 2D segmentation model.

\subsection{Knowledge Distillation}
KD is a powerful technique for transferring insights from one network to another while training constructively. This method is particularly beneficial when dealing with limited labeled data, as it distills knowledge from a well-trained teacher network to a student network. A primary aspect of KD is its focus on reducing disparities between the teacher and student networks during training, often resulting in the student achieving comparable or even superior performance at inference. \citet{hinton2015distilling} introduced the concept of KD for neural networks, transferring knowledge while reducing the difference between two logits. In recent years, several methods \citep{romero2014fitnets,wang2020collaborative,chen2021distilling} have evolved to incorporate feature-based distillation that enables the student network to learn more abundant information from the teacher by transferring knowledge at intermediate layers. \citet{chen2021distilling} discovered that the high-level stage of the student has a great capacity to learn useful information from the teacher's low-level features, such as the edge or texture. Apart from high-level vision tasks (\eg, classification, detection) of traditional KD methods \citep{hinton2015distilling,romero2014fitnets,chen2021distilling}, \citet{wang2020collaborative} proposed a collaborative distillation framework for universal neural style transfer based on the encoder-decoder design and obtained optimal results when distillation was applied solely to the encoder. Moreover, \citet{qin2021efficient} developed a distillation module tailored for medical image segmentation applicable to CNNs that conform with the encoder-decoder architecture.

In this context, the feature-level KD method is utilized for the 7T-absent domain to effectively harness UHF feature representations derived from a teacher network that is trained in the 7T-present domain via a paired 3T-7T dataset. Such a KD strategy enables the student network that typically processes LF images to benefit from the enriched UHF information, thus enhancing its segmentation capabilities without explicit access to 7T images.

\begin{figure}[t!]
\centering
\includegraphics[width=\textwidth]{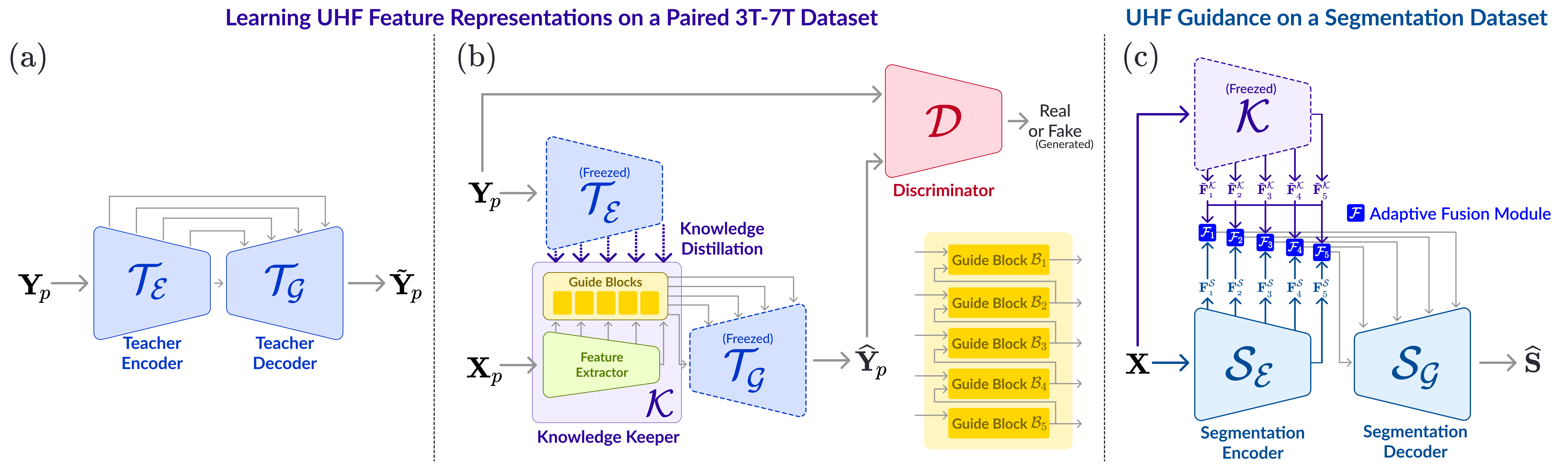}
\caption{Schematic overview of our method. (a) A teacher network $\mathcal{T}$ is trained to extract meaningful UHF representations via 7T image reconstruction. (b) The guide blocks $\mathcal{B}$ within our KKN $\mathcal{K}$ learn how to map a 3T image $\mathbf{X}_p$ into UHF feature representations. (c) The AFMs $\mathcal{F}$ integrate UHF feature representations from the pre-trained KKN $\mathcal{K}$ with LF features $\mathbf{F}^\mathcal{S}_i$ extracted by a segmentation encoder $\mathcal{S}_\mathcal{E}$ and then transfer such fused features to the segmentation decoder $\mathcal{S_G}$ to produce 7T-guided segmentation masks $\mathbf{\widehat{S}}$.}\label{overall_framework}
\end{figure}


\section{Method}
\label{section:method}
The main goal of the proposed method is to learn UHF feature representations on a paired 3T-7T dataset via KD and then provide assimilable feature-level UHF guidance without 7T images for segmenting an LF image (\eg, 3T images). In Figure~\ref{overall_framework}, we illustrate the overall framework of the proposed method, which comprises three core components: a teacher network $\mathcal{T}$, a KKN $\mathcal{K}$, and AFMs $\mathcal{F}$ for an independent segmentation network $\mathcal{S}$ (\ie, baseline models). Given the paired 3T-7T dataset $\{\mathbf{X}_p, \mathbf{Y}_p\}$ of a 7T-present domain, we first train the teacher network $\mathcal{T}$ on high-quality 7T MRI data $\mathbf{Y}_p$ to capture the meaningful UHF feature representation. Contrary to the teacher network $\mathcal{T}$, KKN $\mathcal{K}$ is trained on 3T MRI data $\mathbf{X}_p$ only but aims to implicitly imitate the 7T characteristics (UHF feature representations) derived from the pre-trained teacher encoder $\mathcal{T_E}$ via guide blocks. In this way, the KKN effectively bridges the gap of properties between LF and UHF imaging by learning to properly transform the informative representations of 7T images using only 3T images. Subsequently, we employ the pre-trained KKN with AFMs $\mathcal{F}$ to aggregate and assimilate the UHF feature representations into segmentation models $\mathcal{S}$. The AFMs serve as the integration point where the UHF-like features $\tilde{\mathbf{F}}^\mathcal{K}$ generated by the KKN are fused with the LF features $\mathbf{F}^\mathcal{S}$ from the independent segmentation model. Therefore, our AFMs ensure that the enhanced contrast and detail of UHF imaging are effectively incorporated into the LF image segmentation process, leading to more precise and detailed segmentation results.


\begin{figure}[t!]
\centering
\includegraphics[width=.82\textwidth]{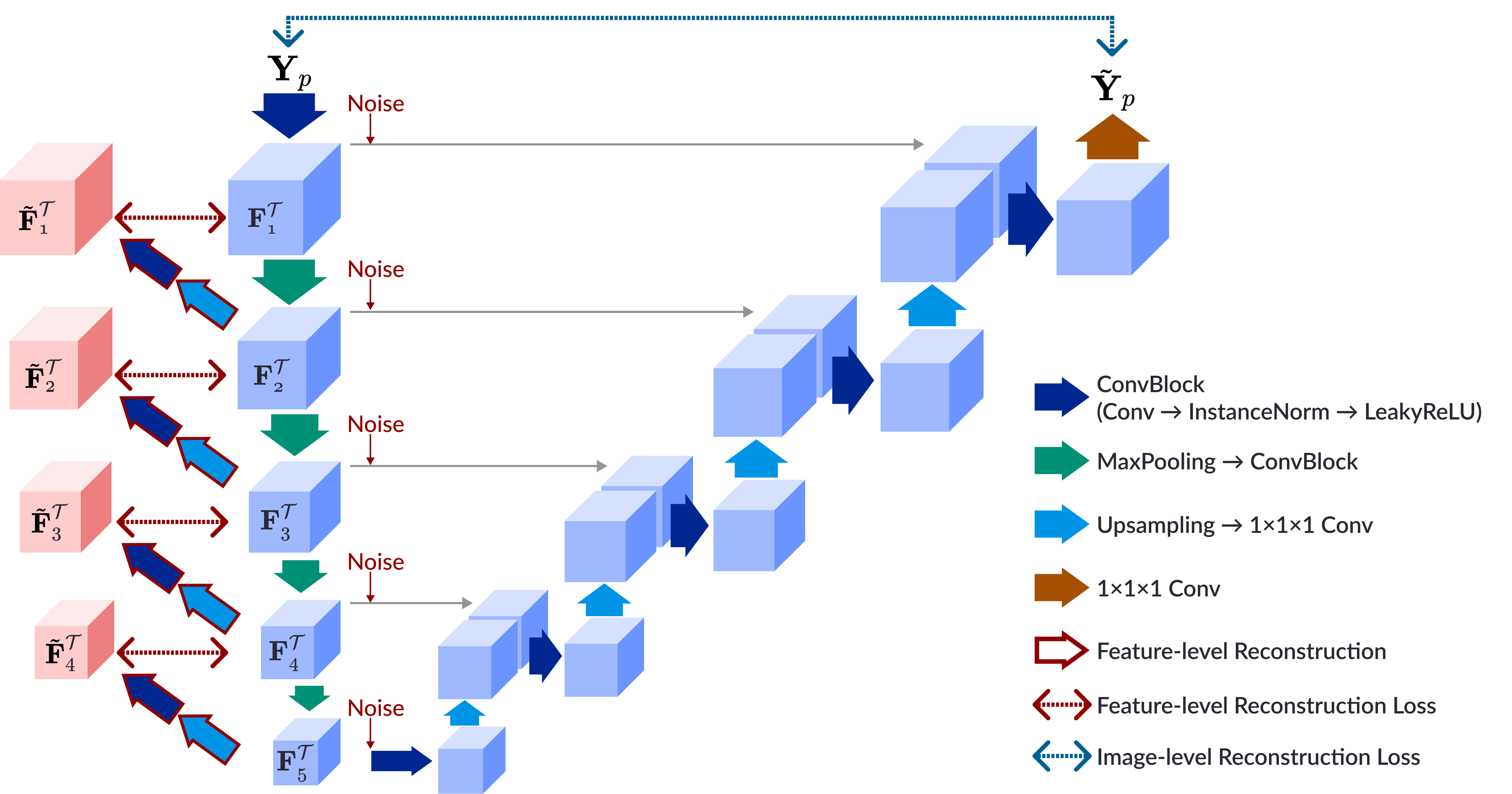}
\caption{Detailed architecture design of the teacher network.}\label{fig:teacher_network}
\end{figure}

To accommodate the various approaches in segmentation models, namely, volume-/slice-based methods, our strategy embraces both 3D and 2D training paradigms to plug the KKN into any given segmentation network. Specifically, in the 3D approach, we train the teacher network and KKN by utilizing entire brain volumes, thereby capturing comprehensive spatial information. Conversely, in the 2D approach, the networks are equipped with three separate 2D CNNs, each corresponding to a different anatomical plane of a volume image: coronal, sagittal, and axial. To synthesize a 3D representation from these 2D networks, we stack the slices along their respective planes, with each slice carrying its predicted values. These slices are then merged, with voxel-level predictions averaged to produce a unified volumetric outcome. This flexibility in selecting either a 3D or 2D KKN according to the specific requirements of the segmentation network allows the optimal transfer of UHF feature representations in scenarios where 7T images are unavailable. In the following subsections, we focus on delineating the network architecture within the context of the 3D training strategy for clarity and coherence.

\subsection{Learning UHF Feature Representations}\label{sec:uhf-learning}
To initiate the KD process, we first train the teacher network $\mathcal{T}$ to extract informative UHF features from a 7T image $\mathbf{Y}_p$ by reconstructing an image $\mathbf{\tilde{Y}}_p$. As depicted in Figure~\ref{fig:teacher_network}, the architecture of our teacher network is a variant of U-Net that includes replacing transposed convolutions with a combination of upsampling and $1 \times 1 \times 1$ convolution. The network structure comprises convolution blocks, each consisting of a $3 \times 3 \times 3$ convolutional layer, followed by instance normalization and the $\operatorname{Leaky ReLU}$ activation function. Furthermore, we use max-pooling with a stride of 2 on a contracting path and tri-linear upsampling followed by a $1 \times 1 \times 1$ convolutional layer on an expanding path.

Given that the skip connections in the network intensely guide the reconstruction of the input image, particularly at lower feature levels\footnote{Low-level features refer to the basic attributes of an image that are extracted in the initial layers of the network, including edges and textures. By contrast, high-level features refer to the complex and abstract attributes extracted in the network's deeper layers, including partial structure and object appearance.}, there is a risk that the network might be hindered from optimally extracting knowledge for distillation. To address this issue and ensure effective training of the teacher network, we apply two principal regularization techniques. As the first regularization, we inject random Gaussian noise $\epsilon=\mathcal{N}(0, \sigma^2)$ into feature maps $\{\mathbf{F}^\mathcal{T}_i\}^{L}_{i=1}=\mathcal{T}_\mathcal{E}(\mathbf{Y}_p)$ when passed to their respective expanding levels via skip connections as
\begin{equation}\label{eq:t_recon}
    \tilde{\mathbf{Y}}_p=\mathcal{T}_\mathcal{G}\left(\mathbf{F}^\mathcal{T}_1 + \epsilon_1,...,\mathbf{F}^\mathcal{T}_L + \epsilon_L\right),
\end{equation}
where $L$ is the total number of encoder blocks. This added noise challenges the teacher network to extract UHF features robustly and reconstruct the image despite slight distortions in terms of visual quality. The second regularization is feature-level reconstruction, which is inspired by multiscale autoencoders \citep{he2021autoencoder}. This is similar to image-level reconstruction, where the entire image is reconstructed via contracting and expanding paths. In feature-level reconstruction, we aim to reproduce $i$-th feature maps $\tilde{\mathbf{F}}^\mathcal{T}_{i}$ from contracted ($i+1$)-th feature maps $\mathbf{F}^\mathcal{T}_{i+1}$. Specifically, such regularization is applied at each encoder block by learning additional expanding modules that consist of upsampling followed by the convolution block (ConvBlock) (refer to Figure~\ref{fig:teacher_network}). The teacher network is thus trained by minimizing an objective function comprising the standard image-level and feature-level reconstruction errors as follows:

\begin{equation}\label{eq:t_loss}
\mathcal{L}_{\text{teacher}} = \lambda_{\text{img}} \|\mathbf{Y}_p - \tilde{\mathbf{Y}}_p\|_{1} + \sum_{i=1}^{L-1} \|\mathbf{F}^{\mathcal{T}}_{i} -  \tilde{\mathbf{F}}^{\mathcal{T}}_{i}\|_{2},
\end{equation}
where $\mathbf{F}^{\mathcal{T}}_{i}$ denotes the output feature maps of the $i$-th block in the teacher encoder $\mathcal{T_E}$ and $\tilde{\mathbf{F}}^{\mathcal{T}}_{i}$ denotes the reconstructed feature maps from $\mathbf{F}^{\mathcal{T}}_{i+1}$. $\lambda_{\text{img}}$ is a weighting hyperparameter. With such regularization techniques, we establish a well-trained teacher network $\mathcal{T}$ for effective KD of the next stage.

\begin{figure}[t!]
\centering
\includegraphics[width=1\textwidth]{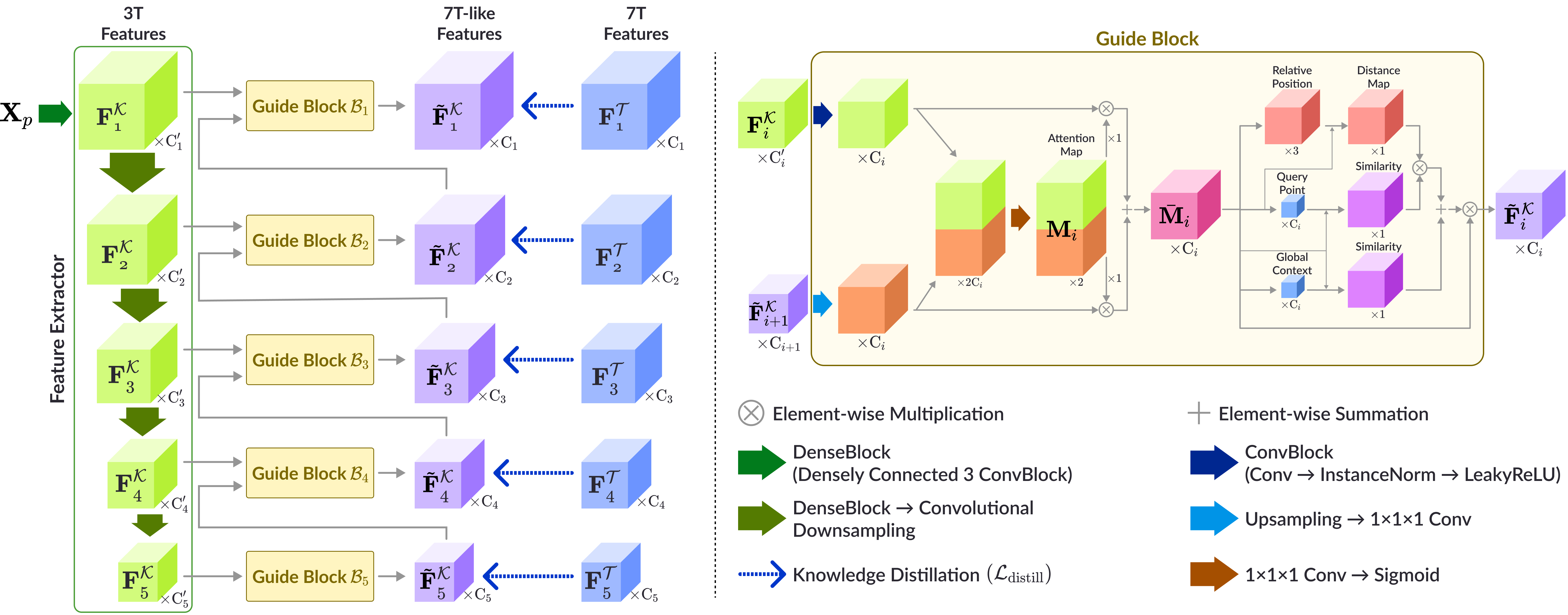}
\caption{Detailed illustration of the knowledge keeper network (KKN) and guide block.}\label{fig:keeper_network}
\end{figure}

\subsection{Effective KD via KKN}\label{sec:kkn-learning}
The KKN plays an intermediate role between two distinct domains: 7T-present and 7T-absent. In the 7T-present domain, where 7T images are available, the KKN learns how to transform a 3T image into UHF feature representations in a task of 7T-like image synthesis. By contrast, in the 7T-absent domain, the pre-trained KKN provides essential UHF information to the segmentation model without the need for 7T images. To achieve this, the KKN includes a feature extractor and several guide blocks $\mathcal{B}$ as detailed in Figures~\ref{overall_framework}(b) and \ref{fig:keeper_network}. The feature extractor is constructed by using densely connected convolution blocks with seven kernel sizes, which are smaller yet deeper than those in the teacher encoder $\mathcal{T_E}$, while guide blocks are devised by combining attention and position-aware recalibration mechanisms that allow the model to transform the LF features into UHF-like features seamlessly.

With the 3T features $\{\mathbf{F}^\mathcal{K}_i\}^L_{i=1}$ extracted from the feature extractor using the 3T image $\mathbf{X}_p$, we first produce the highest-level 7T-like feature $\tilde{\mathbf{F}}^\mathcal{K}_L$ by exploiting the highest-level 3T feature $\mathbf{F}^\mathcal{K}_L$ and their corresponding guide block $\mathcal{B}_L$ at the last layer. The 7T-like feature $\tilde{\mathbf{F}}^\mathcal{K}_L$ is subsequently fed into the guide block $\mathcal{B}_{L-1}$ along with the lower-level 3T feature ${\mathbf{F}}^\mathcal{K}_{L-1}$ as $\tilde{\mathbf{F}}^\mathcal{K}_{L-1} = \mathcal{B}_{L-1}(\tilde{\mathbf{F}}^\mathcal{K}_L, \mathbf{F}^\mathcal{K}_{L-1})$. Without loss of generality, we use the $i$-th guide block as an example (refer to the right side of Figure~\ref{fig:keeper_network}). In the guide block, 7T-like features are upscaled coherently to the 3T features to standardize their dimensions of both channels and spatial resolution, and then the attention map $\mathbf{M}_i$ is created by conducting the concatenation operation followed by $1\times 1\times 1$ convolution ($\operatorname{Conv3D}_{1\times1\times1}$) with a sigmoid function $g$ as 
\begin{equation}\label{eq:att_map}
\mathbf{M}_{i}:=g\left(\operatorname{Conv3D}_{1\times1\times1}\left(\mathbf{F}^\mathcal{K}_{i}\odot\operatorname{Conv3D}_{1\times1\times1}\left(\operatorname{INTRPL}\left(\tilde{\mathbf{F}}^\mathcal{K}_{i+1}\right)\right)\right)\right), 
\end{equation}
where $\odot$ and $\operatorname{INTRPL}$ denote the concatenation and tri-linear interpolation, respectively. Given such an attention map $\mathbf{M}_i$, each of the 3T and upscaled 7T features is revamped via the attention map that highlights specific regions of interest so that it enhances the relevant features while diminishing less important ones:
\begin{equation}
    \bar{\mathbf{M}}_i=\left(\mathbf{M}_i[0, :] \otimes \mathbf{F}^\mathcal{K}_i\right) + \left(\mathbf{M}_i[1, :] \otimes \operatorname{INTRPL}\left(\tilde{\mathbf{F}}^\mathcal{K}_{i+1}\right)\right),
\end{equation}
where $\otimes$ denotes element-wise multiplication. This map $\bar{\mathbf{M}}_i$ is transferred to the position-aware recalibration module (PRM) \citep{ma2020position} to yield 7T-like features by leveraging the relative positions and similarities between the global context and the most salient feature as $\tilde{\mathbf{F}}^\mathcal{K}_{i}=\operatorname{PRM}(\bar{\mathbf{M}}_i)$. By circularly fulfilling up to the lowest-level guide block $\mathcal{B}_1$, we eventually acquire 7T-like features $\{\tilde{\mathbf{F}}^\mathcal{K}_i\}^L_{i=1}$ across all levels of the network. The guide block used in the last layer omits the attention operation and only performs PRM because it is inevitably infeasible to obtain the 7T-like feature in the highest level for calculating the attention map $\mathbf{M}_i$ in Eq.~\ref{eq:att_map}.

For training our KKN $\mathcal{K}$, we exploited both a pre-trained teacher network $\mathcal{T}$ and a patch discriminator $\mathcal{D}$ \citep{isola2017image} for 7T-like image synthesis. To begin with, the feature maps of the KKN were optimized to match UHF features that were extracted from a 7T image $\mathbf{Y}_p$ using the teacher encoder $\mathcal{T_E}$. The loss for KD is defined as follows:
\begin{equation}\label{eq:distill_loss}
\mathcal{L}_{\text{distill}} = \sum_{i=1}^{L} (\|\mathbf{F}^{\mathcal{T}}_{i} - \tilde{\mathbf{F}}^{\mathcal{K}}_{i}\|_{2} + \|\mu(\mathbf{F}^{\mathcal{T}}_{i}) - \mu(\tilde{\mathbf{F}}^{\mathcal{K}}_{i})\|_{2} + \|\sigma(\mathbf{F}^{\mathcal{T}}_{i}) - \sigma(\tilde{\mathbf{F}}^{\mathcal{K}}_{i})\|_{2}),
\end{equation}
where $\mathbf{F}^{\mathcal{T}}_{i}$ denotes the feature maps of the $i$-th block in the teacher encoder $\mathcal{T_E}$. The channel-wise mean $\mu$ and standard deviation $\sigma$ are measured to match the distribution of feature maps \citep{wang2021imagine} between the KKN $\mathcal{K}$ and teacher encoder $\mathcal{T_E}$. Using feature maps of the KKN $\mathcal{K}$, the teacher decoder $\mathcal{T_G}$ generates the 7T-like image $\mathbf{\widehat{Y}}_p$ (refer to Figure~\ref{overall_framework}(b)) and enables the computation of the synthesis loss $\mathcal{L}_{\text{syn}} = \| \mathbf{Y}_p - \mathbf{\widehat{Y}}_p \|_{1}$. For realistic 7T-like image synthesis, our KKN $\mathcal{K}$ is adversarially trained with the discriminator $\mathcal{D}$ using a loss of $\mathcal{L}_{\text{adv}} = \| \mathcal{D}(\mathbf{\widehat{Y}}_p) - \boldsymbol{1} \|_{2}$, where $\boldsymbol{1}$ denotes a constant value for the real label. In this way, the KKN effectively emulates the functionality of the teacher encoder $\mathcal{T_E}$, allowing it to replicate the process of transforming features of 3T images into 7T-like features. Consequently, the comprehensively objective function for the KKN is formulated by integrating distillation loss, synthesis loss, and adversarial loss \citep{mao2017least} as follows:
\begin{equation}\label{eq:keeper_loss}
\mathcal{L}_{\text{keeper}} = \mathcal{L}_{\text{distill}} + \lambda_{\text{syn}}\mathcal{L}_{\text{syn}} + \lambda_{\text{adv}}\mathcal{L}_{\text{adv}},
\end{equation}
where $\lambda_{\text{syn}}$ and $\lambda_{\text{adv}}$ are hyperparameters for the magnitudes of weighting. By optimizing the total loss $\mathcal{L}_{\text{keeper}}$, our KKN $\mathcal{K}$ is able to learn UHF feature representations, which are available for the 7T-absent domain.

\begin{figure}[t!]
\centering
\includegraphics[width=\textwidth]{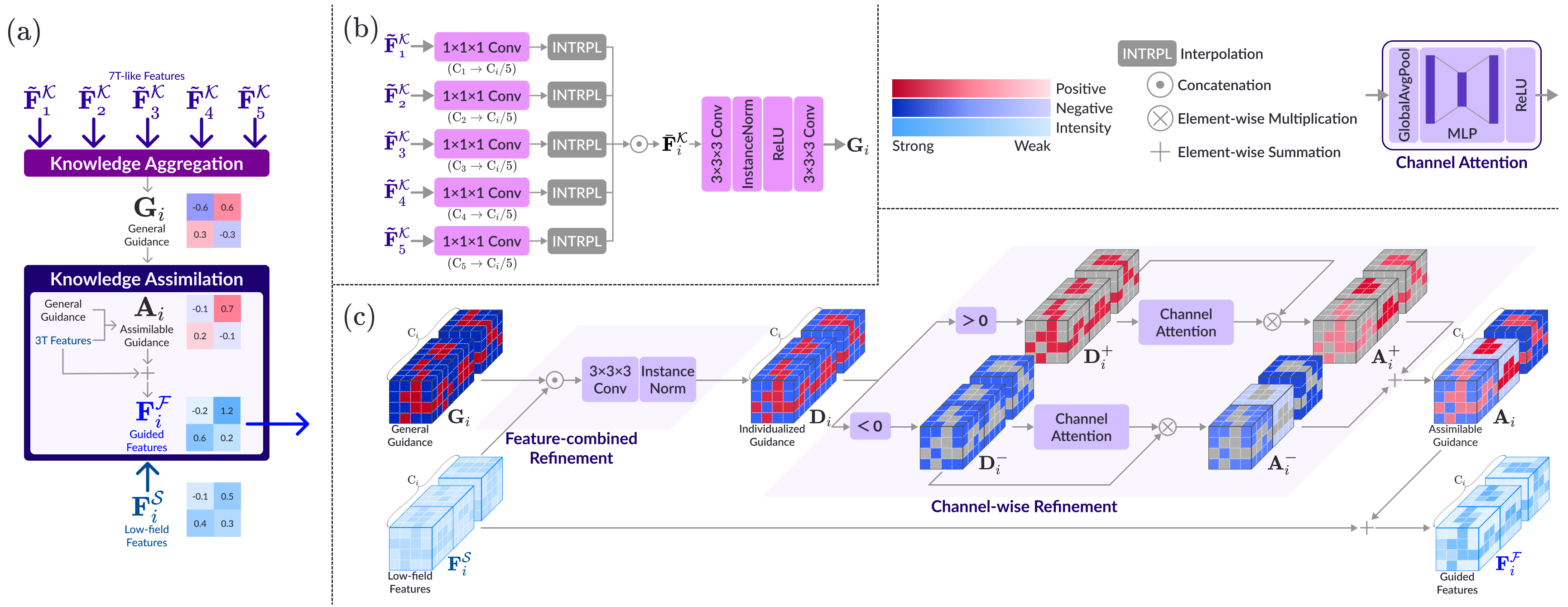}
\caption{(a) Knowledge flow in the $i$-th level adaptive fusion module (AFM). (b) In knowledge aggregation, 7T-like features are converted into general UHF guidance that has strong contrast properties. (c) In knowledge assimilation, the general guidance is refined to be individualized and assimilated to adjust the intensity values of LF features.}\label{fusion_module}
\end{figure}

\subsection{UHF Guidance for LF-Based Brain Segmentation}\label{sec:lf-based-seg-training}
Our goal of brain segmentation is to reveal intricate and detailed structures that are typically obscured when relying solely on LF image features. 
To leverage the potential of UHF feature representations, the pre-trained KKN $\mathcal{K}$ transforms the LF image $\mathbf{X}$ into 7T-like features across all levels $\{\tilde{\mathbf{F}}^\mathcal{K}_1,...,\tilde{\mathbf{F}}^\mathcal{K}_L\}$ and conveys them to the AFMs $\mathcal{F}$ at corresponding levels as indicated by the purple lines in Figure~\ref{overall_framework}(c). Concurrently, inherent LF image features $\{\mathbf{F}^\mathcal{S}_1,...,\mathbf{F}^\mathcal{S}_L\}$ are extracted by the segmentation encoder $\mathcal{S_E}$ and fed into the AFM as denoted by the blue lines in Figure~\ref{overall_framework}(c).

In the $i$-th AFM $\mathcal{F}_i$, 7T-like features are injected as new knowledge in the 7T-absent domain and altered to the optimal UHF guidance through a knowledge flow that encompasses both knowledge aggregation and assimilation as shown in Figure~\ref{fusion_module}(a). In particular, our approach tailors UHF guidance to have both positive and negative values, enabling specific voxels to be conspicuous by being either brighter or darker relative to their surroundings. Adding such assimilable guidance to LF features facilitates the highlighting of structural details that may otherwise remain concealed in LF images owing to low contrast while maintaining the inherent appearance information.

The process of generating assimilable guidance $\mathbf{A}_i$ for the LF features of each level begins with the creation of the comprehensive guidance map, that is, general guidance $\mathbf{G}_i$ as shown in Figure~\ref{fusion_module}(b). Given all levels of 7T-like features $\{\tilde{\mathbf{F}}^\mathcal{K}_1,...,\tilde{\mathbf{F}}^\mathcal{K}_L\}$, we first carry out the knowledge aggregation that employs $\operatorname{INTRPL}$ followed by $\operatorname{Conv3D}_{1\times1\times1}$ to standardize their dimensions, both in terms of channels and spatial resolution. By merging these adjusted coarse-to-fine features, we collect abundant UHF information to form a representative 7T feature descriptor $\bar{\mathbf{F}}^\mathcal{K}_i$ for the $i$-th level guidance as

\begin{equation}
    \bar{\mathbf{F}}^\mathcal{K}_i := \bigg\Vert_{i=1}^L\left(\operatorname{INTRPL}\left(\operatorname{Conv3D}_{1\times1\times1}\left(\tilde{\mathbf{F}}_{i}^\mathcal{K}\right)\right)\right),
\end{equation}
where $||$ and $L$ indicate the operator of concatenation and total encoder blocks, respectively. The 7T feature descriptor $\bar{\mathbf{F}}^\mathcal{K}_i$ includes a broad spectrum of values and pronounced contrast information from knowledge aggregation. To derive general guidance $\mathbf{G}_i$, the 7T feature descriptor is then processed by an additional convolution block that comprises a convolution operation with a trainable $3\times 3\times 3$ kernel ($\operatorname{Conv3D}_{3\times 3\times 3}$), instance normalization, and $\operatorname{ReLU}$ activation function, followed by $\operatorname{Conv3D}_{3\times 3\times 3}$ without an activation function. In this way, we obtain the general guidance $\mathbf{G}_i$, which assigns linearity with an unlimited range of values in the 7T feature descriptor.

Subsequently, as depicted in Figure~\ref{fusion_module}(c), this general guidance $\mathbf{G}_i$ is suitably adapted to be more assimilable for LF features $\mathbf{F}^\mathcal{S}_{i}$ via two refinement processes: feature-combined and channel-wise refinement. This refinement entails comparing the general UHF guidance with the $i$-th level LF features to identify regions affected by excessive alterations in global contrast. We then determine the degree to which intensity values should be adjusted in each channel, thereby establishing the optimal degree of UHF guidance intervention in the LF features. In feature-combined refinement, the individualized guidance $\mathbf{D}_{i}$ for each level is formulated by taking into account both the general guidance and the LF features as
\begin{equation}
\mathbf{D}_{i} = \text{IN}(\operatorname{Conv3D}_{3\times3\times3}(\mathbf{G}_{i}\odot\mathbf{F}^{\mathcal{S}}_{i})),
\end{equation}
where $\odot$ denotes the concatenation and IN denotes an instance normalization. 
The process of refining individualized guidance $\mathbf{D}_{i}$ is further elaborated by channel-wise refinement, which establishes a dual-path method for guiding the directionality of intensity adjustments. In particular, this refinement involves dividing the individualized guidance $\mathbf{D}_{i}$ into two separate components according to their sign, resulting in positive ($\mathbf{D}^{+}_{i} > 0$) and negative ($\mathbf{D}^{-}_{i} < 0$) guidance. This division allows a more nuanced refinement by clearly distinguishing whether the intensity values in a given region are to be increased or decreased. Subsequent to the guidance division, we acquire the individualized channel descriptors $\mathbf{d}^{*}_{i}$ by estimating the summary statistics of a channel via global average pooling (GAP) as $\mathbf{d}^{*}_{i} = \operatorname{GAP}(\mathbf{D}^{*}_{i})$. For brevity, we have denoted the sign including both positive ($+$) and negative ($-$) as the $*$ symbol, \ie, $\mathbf{d}_i^*=\mathbf{d}_i^{+/-}$ and $\mathbf{D}_i^*=\mathbf{D}_i^{+/-}$. 

Each individualized channel descriptor $\mathbf{d}^{*}_{i}$ is assimilated by using a squeeze and excitation \citep{hu2018squeeze}, functioning as a channel attention mechanism. Contrary to the conventional sigmoid activation function, which typically allocates a probability, we apply a ReLU activation function because it measures the magnitude of guidance for each channel without suppressing weights, thus ensuring that the full intensity range is available for use. Accordingly, the assimilable guidance $\mathbf{A}_i^{*}$ is produced by performing the multiplication with corresponding individualized guidance $\mathbf{D}_{i}^{*}$ as
\begin{equation}
\mathbf{A}^{*}_{i} = \operatorname{ReLU}(W_{1}^{*} \otimes \operatorname{ReLU} (W_{0}^{*} \otimes \mathbf{d}^{*}_{i})) \otimes \mathbf{D}^{*}_{i}, 
\end{equation}
where $W_{0} \in \mathbb{R}^{\frac{1}{2}C_{i} \times C_{i}}$ and $W_{1} \in \mathbb{R}^{C_{i} \times \frac{1}{2}C_{i}}$ are trainable embedding weights and $\otimes$ denotes element-wise multiplication. These two distinct assimilable guidance types ($\mathbf{A}_i^+$ and $\mathbf{A}_i^-$) are reassembled into representative assimilable guidance, expressed as $\mathbf{A}_{i}=\mathbf{A}^{+}_{i}+\mathbf{A}^{-}_{i}$. By adding assimilable guidance $\mathbf{A}_{i}$ to LF features $\mathbf{F}^{\mathcal{S}}_i$ using element-wise summation, we can provide the segmentation decoder $\mathcal{S_G}$ with guided features $\mathbf{F}^{\mathcal{F}}_i=\mathbf{A}_{i}+\mathbf{F}^{\mathcal{S}}_i$ that exhibit a more pronounced structural appearance and clearly enhanced contrast:
\begin{equation}
\widehat{\mathbf{S}}=\mathcal{S_G}(\mathbf{F}^{\mathcal{F}}_1, ..., \mathbf{F}^{\mathcal{F}}_L),
\end{equation}
where $\widehat{\mathbf{S}}$ indicates the predicted segmentation mask. The segmentation model $\mathcal{S}$ applying our AFM was trained with identical objective functions that were used in each baseline model $\mathcal{S}$. The PyTorch-like pseudo code for the overall training phases has been summarized in Algorithm~\ref{pseudocode}.

\begin{algorithm}[!t]
\footnotesize
\caption{Algorithm of our overall proposed framework}
    \begin{algorithmic}[1]\label{pseudocode}
    \REQUIRE{Paired 3T-7T dataset $\{\mathbf{X}_p, \mathbf{Y}_p\}$, LF image $\mathbf{X}$, Teacher encoder $\mathcal{T_E}$, Teacher decoder $\mathcal{T_G}$, knowledge keeper (KKN) $\mathcal{K}$, Discriminator $\mathcal{D}$, adaptive fusion module (AFM) $\mathcal{F}$, Segmentation encoder $\mathcal{S_E}$, Segmentation decoder $\mathcal{S_G}$, Model's tunable parameters $\theta_*$, Total number of layers $L$, Hyperparameters $\lambda_*$, Learning rate $\eta$.}

    \textbf{\textit{Phase 1. Learning of UHF Feature Representations} (Section~\ref{sec:uhf-learning})}
    \FOR{$epoch = 1, 2,\dots$}
    \STATE Extract the 7T features: $\{\mathbf{F}^\mathcal{T}_i\}^{L}_{i=1}=\mathcal{T}_\mathcal{E}(\mathbf{Y}_p)$
    \STATE Reconstruct the 7T image: $\tilde{\mathbf{Y}}_p=\mathcal{T_G}(\mathbf{F}^\mathcal{T}_1,...,\mathbf{F}^\mathcal{T}_L)$ 
    \STATE Calculate $\mathcal{L}_{\text{teacher}} = \lambda_{\text{img}} \|\mathbf{Y}_p - \tilde{\mathbf{Y}}_p\|_{1} + \sum_{i=1}^{L-1} \|\mathbf{F}^{\mathcal{T}}_{i} -  \tilde{\mathbf{F}}^{\mathcal{T}}_{i}\|_{2}$ 
    \STATE Update the teacher encoder $\mathcal{T_E}$ and the decoder $\mathcal{T_G}$:
    
    \STATE $\theta_{\mathcal{T_{E,G}}} \leftarrow \theta_{\mathcal{T_{E,G}}} - \eta\nabla_{\theta_{\mathcal{T_{E,G}}}}\mathcal{L}_\text{teacher}$
    \ENDFOR
    \RETURN{Well-trained teacher encoder $\mathcal{T_E}$ and decoder $\mathcal{T_G}$}
    
    \textbf{\textit{Phase 2. Effective Knowledge Distillation via KKN} (Section~\ref{sec:kkn-learning})}
    \FOR{$epoch = 1, 2,\dots$}
    \STATE{\em // Using the pre-trained teacher encoder $\mathcal{T_E}$ and decoder $\mathcal{T_G}$}
    \STATE Extract the 7T features from teacher encoder: $\{\mathbf{F}^\mathcal{T}_i\}^{L}_{i=1}=\mathcal{T}_\mathcal{E}(\mathbf{Y}_p)$
    \STATE Extract the 7T-like features from KKN: $\{\tilde{\mathbf{F}}^\mathcal{K}_i\}^L_{i=1}=\mathcal{K}(\mathbf{X}_p)$
    \STATE Generate the 7T-like image: $\widehat{\mathbf{Y}}_p=\mathcal{T_G}(\tilde{\mathbf{F}}^\mathcal{K}_1, ..., \tilde{\mathbf{F}}^\mathcal{K}_L)$
    \STATE Calculate for Eq. \eqref{eq:keeper_loss} as:
    \STATE $\mathcal{L}_{\text{distill}} = \sum_{i=1}^{L} (\|\mathbf{F}^{\mathcal{T}}_{i} - \tilde{\mathbf{F}}^{\mathcal{K}}_{i}\|_{2} + \|\mu(\mathbf{F}^{\mathcal{T}}_{i}) - \mu(\tilde{\mathbf{F}}^{\mathcal{K}}_{i})\|_{2} + \|\sigma(\mathbf{F}^{\mathcal{T}}_{i}) - \sigma(\tilde{\mathbf{F}}^{\mathcal{K}}_{i})\|_{2})$ 
    \STATE $\mathcal{L}_{\text{syn}} = \| \mathbf{Y}_p - \mathbf{\widehat{Y}}_p \|_{1}$
    \STATE $\mathcal{L}_{\text{adv}} = \| \mathcal{D}(\mathbf{\widehat{Y}}_p) - \boldsymbol{1} \|_{2},$ \text{ where $\boldsymbol{1}$ is a real label}
    \STATE Update the KKN $\mathcal{K}$:
    \STATE $\theta_{\mathcal{K}} \leftarrow \theta_{\mathcal{K}} - \eta\nabla_{\theta_{\mathcal{K}}}(\mathcal{L}_{\text{distill}} + \lambda_{\text{syn}}\mathcal{L}_{\text{syn}} + \lambda_{\text{adv}}\mathcal{L}_{\text{adv}}$) 
    \ENDFOR 
    \RETURN{Well-trained KKN $\mathcal{K}$}
    
    \textbf{\textit{Phase 3. UHF Guidance for LF-Based Brain Segmentation} (Section~\ref{sec:lf-based-seg-training})}
    \FOR{$epoch = 1, 2,\dots$}
    \STATE{\em // Using the pre-trained KKN $\mathcal{K}$}
    \STATE Extract the UHF-like features from KKN: $\{\tilde{\mathbf{F}}^\mathcal{K}_i\}^L_{i=1}=\mathcal{K}(\mathbf{X})$
    \STATE Extract the LF features from segmentation encoder: $\{\mathbf{F}^\mathcal{S}_i\}^L_{i=1}=\mathcal{S_E}(\mathbf{X})$

    \STATE Generate guided features from AFM: $\{{\mathbf{F}}^\mathcal{F}_i\}^L_{i=1}=\mathcal{F}(\tilde{\mathbf{F}}^\mathcal{K}, \mathbf{F}^\mathcal{S})$
    \STATE Produce the segmentation mask: $\widehat{\mathbf{S}}=\mathcal{S_G}(\mathbf{F}^{\mathcal{F}}_1, ..., \mathbf{F}^{\mathcal{F}}_L)$
    \STATE Update $\theta_{\mathcal{S}_{\mathcal{E,G}}}$ and $\theta_{\mathcal{F}}$ via objective functions used in each segmentation model $\mathcal{S}$
    \ENDFOR
    \end{algorithmic}
\end{algorithm}

\section{Experiments}
\label{section:exp}
\subsection{Datasets}
\subsubsection{Paired 3T-7T Dataset}
We used 15 pairs of 3T and 7T T1-weighted MR images scanned from 15 adults, and such images were acquired using Siemens Magnetom Trio 3T and 7T MRI scanners. For 3T images, 224 coronal slices were obtained with repetition time (TR) = 1900 $ms$, echo time (TE) = 2.16 $ms$, inversion time (TI) = 900 $ms$, flip angle (FA) = 9$^{\circ}$, and voxel size = $1.0~mm \times 1.0~mm \times 1.0~mm$. For 7T images, 191 sagittal slices were obtained with repetition time (TR) = 6000 $ms$, echo time (TE) = 2.95 $ms$, inversion time (TI) = 800/2700 $ms$, flip angle (FA) = 4$^{\circ}$/4$^{\circ}$, and voxel size = $0.65~mm \times 0.65~mm \times 0.65~mm$.

Regarding the input processing for training the 3D network, we inevitably cropped the background that did not include anatomical structures of the brain owing to the immense memory issue\footnote{Without background cropping, our model could have faced operational challenges (or perhaps even failed to train) because the batch size for 3D network training was initialized to 1 as reported in Section \ref{subsec: network-implementation}.}. Similar to the previous 3T-to-7T translation methods \citep{zhang2018dual,qu2020synthesized}, we normalized the input intensity to the range of [0, 1] and applied the histogram-matching technique to the 3T and 7T images. In the experiments, we adopted a leave-one-out cross-validation strategy to maximize data utilization. Specifically, one pair of 3T and 7T images was used for testing, and the remaining pairs were used for training. For the validation of segmentation performance in the 7T-present domain, we obtained the ground truth over the tissue-segmented masks from the 7T image using ANTs \citep{avants2011open}.

\subsubsection{Segmentation Datasets}
To verify the adaptability and scalability of the proposed method, we used two public segmentation datasets: (i) LF images: the Internet Brain Segmentation Repository (IBSR) for brain tissue segmentation and (ii) MALC for whole-brain segmentation.

The IBSR dataset includes MR images of 18 subjects aged 7-71 years and corresponding tissue-segmented masks annotated by an expert. To achieve identical sizes for all volumes and reduce the computational cost, we resampled the MR images to the coherent voxel size as $1.0~mm \times 1.0~mm \times 1.0~mm$ and then cropped such images with the same processing as that used in the 3D network training on the paired 3T-7T dataset. In the experiments, we partitioned the dataset into 7 training, 3 validation, and 8 testing samples according to the specific criteria that each dataset had a similar age population. The MALC dataset contains 35 MR images with manual annotation from 30 subjects, divided into 25 unique subjects and 5 subjects scanned twice. All scans have a coherent voxel size, $1.0~mm \times 1.0~mm \times 1.0~mm$. We identically followed the MALC 2012 challenge setting that assigned 15 samples for training and 20 for testing and used 5 volumes of the training dataset for validation. The labels of manual annotation were remapped from 1 to 27, similar to the mapping strategy of QuickNAT \citep{roy2019quicknat}.

\subsection{Evaluation Metrics}
We adopted two quantitative evaluation metrics to verify the quality of synthesized brain images: peak SNR (PSNR) and structural similarity (SSIM). PSNR provides an objective measure of image quality based on pixel-level differences, which are the reconstruction errors between the original and synthesized images, while SSIM is more sensitive to changes in luminance, contrast, and structure, making it better for detecting subtle distortions. Higher PSNR and SSIM values indicate better quality in terms of pixel-level accuracy and preservation of structural characteristics. The dice similarity coefficient (DSC) is used to evaluate brain segmentation, which quantifies the overlap between the segmented region (\ie, a predicted segment) and the ground truth. In particular, DSC is suitable for assessing segmentation results regardless of the size of the region of interest, such as whole brain and brain tissue segmentation.

\subsection{Implementation Details}\label{subsec: network-implementation} When training the teacher network and KKN, we used Adam optimizer with a fixed learning rate of 0.0002 and random data augmentation, including rotation, scaling, and left-right flipping. The batch size was 1 and 32 for 3D and 2D networks, respectively. The hyperparameters of each objective function, $\lambda_{\text{img}}$, $\lambda_{\text{syn}}$, and $\lambda_{\text{adv}}$, were set to 100, 100, and 0.5, respectively. The segmentation network was initialized with a pre-trained baseline model and then trained with fusion modules using the same strategy as the baseline. We implemented all networks using the PyTorch framework on a workstation equipped with an Intel Xeon Silver4216 CPU @ 2.10 GHz and a single NVIDIA RTX A6000 GPU (48GB memory).

\subsection{Results and Analyses}
\begin{figure}[t!]
\centering
\includegraphics[width=\textwidth]{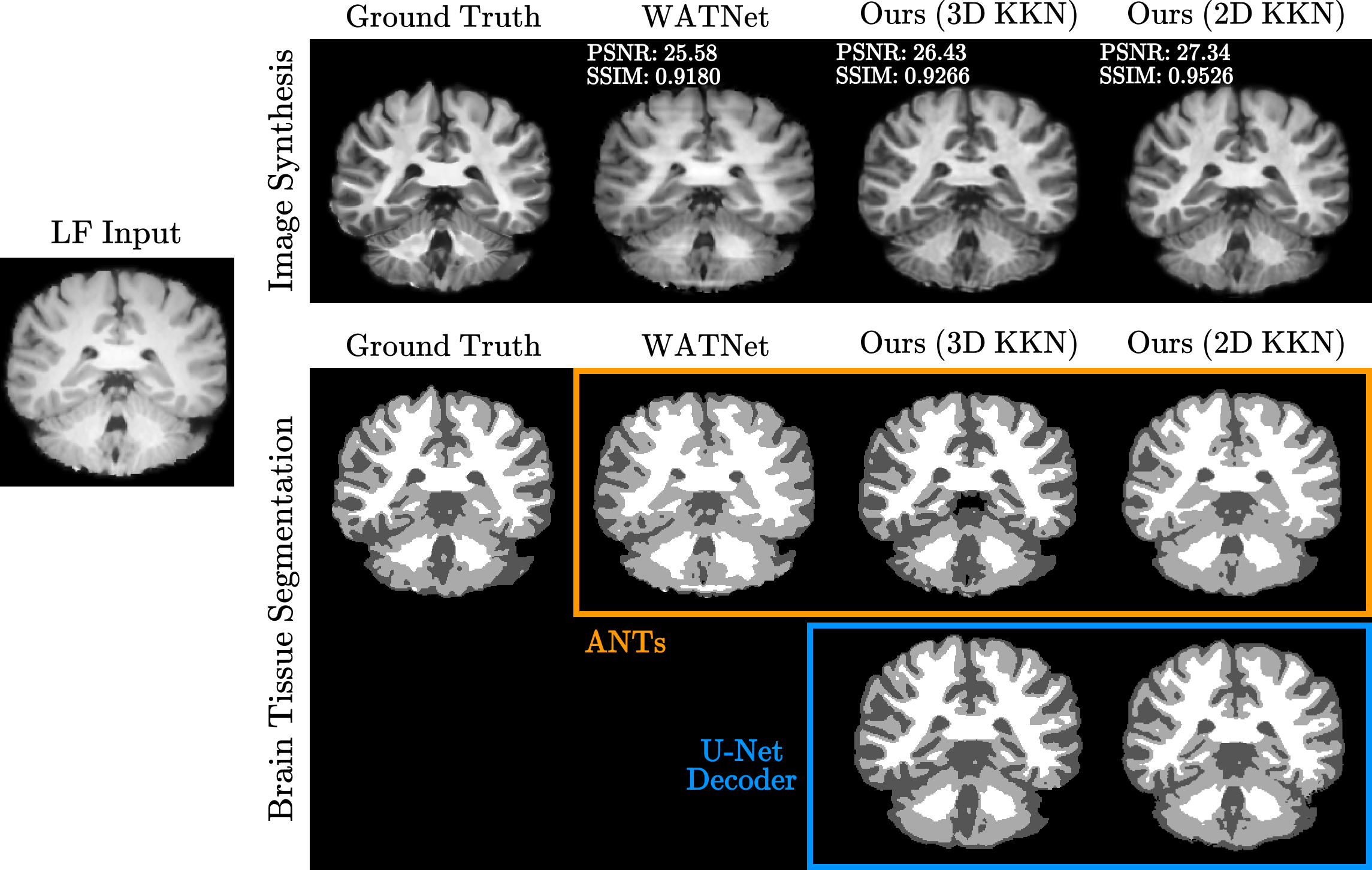}
\caption{Illustration of 7T-like image synthesis (top) and 7T-guided brain tissue segmentation (bottom) results on the paired 3T-7T dataset.}\label{qualitative_comparison_paired}
\end{figure}

\subsubsection{Reliability of UHF Feature Representations} 
We implemented the respective 3D and 2D training strategies on the paired 3T-7T dataset to learn UHF feature representations for both volume- and slice-based approaches. To demonstrate the reliability of UHF feature representations, we performed a qualitative and quantitative evaluation for 7T-like image synthesis and 7T-guided segmentation by comparison with WATNet \citep{qu2020synthesized}, which achieved state-of-the-art performance in 7T image synthesis.

As shown at the top of Figure~\ref{qualitative_comparison_paired}, the results of the proposed method (\ie, 3D KKN and 2D KKN) revealed the superior quality of 7T synthesis compared to the ground truth. Overall, the 7T-like features from our method were proficiently imposed on the 3T input such that we observed distinct contrast differences within the brain, especially in the cerebellum and GM areas. By contrast, WATNet could not adequately eliminate the unique 3T properties despite providing a plausible exhibition regarding appearance preservation. The result of WATNet even showed artifacts such as line noise and blurry effects throughout the brain image, suggesting that UHF feature representations were not reflected seamlessly in the process of 7T synthesis. In addition to qualitative observations, we quantitatively measured the performance using PSNR and SSIM as highlighted in the top-left corner of the respective images in Figure~\ref{qualitative_comparison_paired}. Similar to the qualitative analysis, the PSNR and SSIM of our method achieved better performance than WATNet, and interestingly, our 2D KKN outperformed the 3D KKN in these metrics (\ie, PSNR: +0.91, SSIM: +0.026). We hypothesized that 2D KKN translated MR images slice by slice with less complexity of the spatial relationships, allowing a more minute examination of each slice.

On the basis of these outcomes, we separated the evaluation of brain tissue segmentation into two cases: using 7T-like synthesized images and UHF feature representation learning. In the case of using 7T-like synthesized images, we directly segmented such images (\ie, the results of image synthesis in 
Figure~\ref{qualitative_comparison_paired}) into three distinct regions—GM, WM, and cerebrospinal fluid (CSF)—utilizing ANTs as depicted in the orange box in Figure~\ref{qualitative_comparison_paired}. In the case of UHF feature representation learning, we trained only a U-Net decoder by taking the pre-trained KKN as an encoder to generate 7T-guided segmented masks from the LF image as depicted in the blue box in Figure~\ref{qualitative_comparison_paired}. Most remarkably, the segmentation results of our method for both scenarios were notably enhanced, particularly in areas where WATNet struggled to perceive regions, such as the morphologically subtle boundaries between GM and WM and the intricate structures within the CSF. Moreover, the UHF guidance provided by the KKN encoder significantly improved the delineation of these regions with more accurate anatomical details in the 7T-absent domain, resulting in segmentation masks that closely resembled those obtained from actual 7T images (results in the blue box). To gain insight into UHF feature representations, we further conducted an extensive analysis comparing the intermediate feature maps generated by the KKN with those produced by the teacher network, revealed in \ref{app:verification-7t-like-feature}. From this comparison, we could confirm that the KKN effectively captured and replicated the teacher network's intricate structural patterns and UHF feature representations.

\begin{table}[t!]\centering\scriptsize\setlength{\tabcolsep}{2pt}
    \caption{Dice similarity coefficient (DSC) of brain tissue segmentation on the paired 3T-7T dataset across 15 folds (mean$\pm$standard deviation). The highest scores are in boldface, and the second-highest scores are underlined.}\label{tab_result_paired}
    \begin{tabular}{cccccccc}
        \toprule
        & \multicolumn{2}{c}{\textbf{Method}} & \multicolumn{4}{c}{\textbf{DSC Score ($\%$)}}\\
        \cmidrule(lr){2-3} \cmidrule(lr){4-7}
        & \textbf{3T-to-7T} & \textbf{Segmentation} & \textbf{CSF} & \textbf{GM} & \textbf{WM} & \textbf{Average}\\
        \midrule
        (a) & WATNet \cite{qu2020synthesized} & ANTs & 64.19$\pm$6.45 & 71.85$\pm$6.67 & 80.04$\pm$7.37 & 72.03$\pm$5.54\\
        (b) & Ours (3D KKN) & ANTs & 72.33$\pm$5.81 & 75.05$\pm$6.40 & 83.49$\pm$7.60 & 76.96$\pm$5.49\\
        (c) & Ours (2D KKN) & ANTs & 72.39$\pm$6.31 & \underline{77.80$\pm$7.14} & 84.99$\pm$7.09 & \underline{78.40$\pm$5.97}\\
        (d) & Ours (3D KKN) & 3D U-Net Decoder & \underline{72.66$\pm$4.99} & 76.08$\pm$4.04 & \underline{86.18$\pm$6.62} & 78.31$\pm$4.03\\
        (e) & Ours (2D KKN) & 2D U-Net Decoder & \textbf{76.50}$\pm$\textbf{4.29} & \textbf{80.55}$\pm$\textbf{6.71} & \textbf{86.33}$\pm$\textbf{7.65} & \textbf{81.13}$\pm$\textbf{5.66}\\
        \bottomrule
    \end{tabular}
\end{table}

\begin{figure}[t!]
\centering
\includegraphics[width=\textwidth]{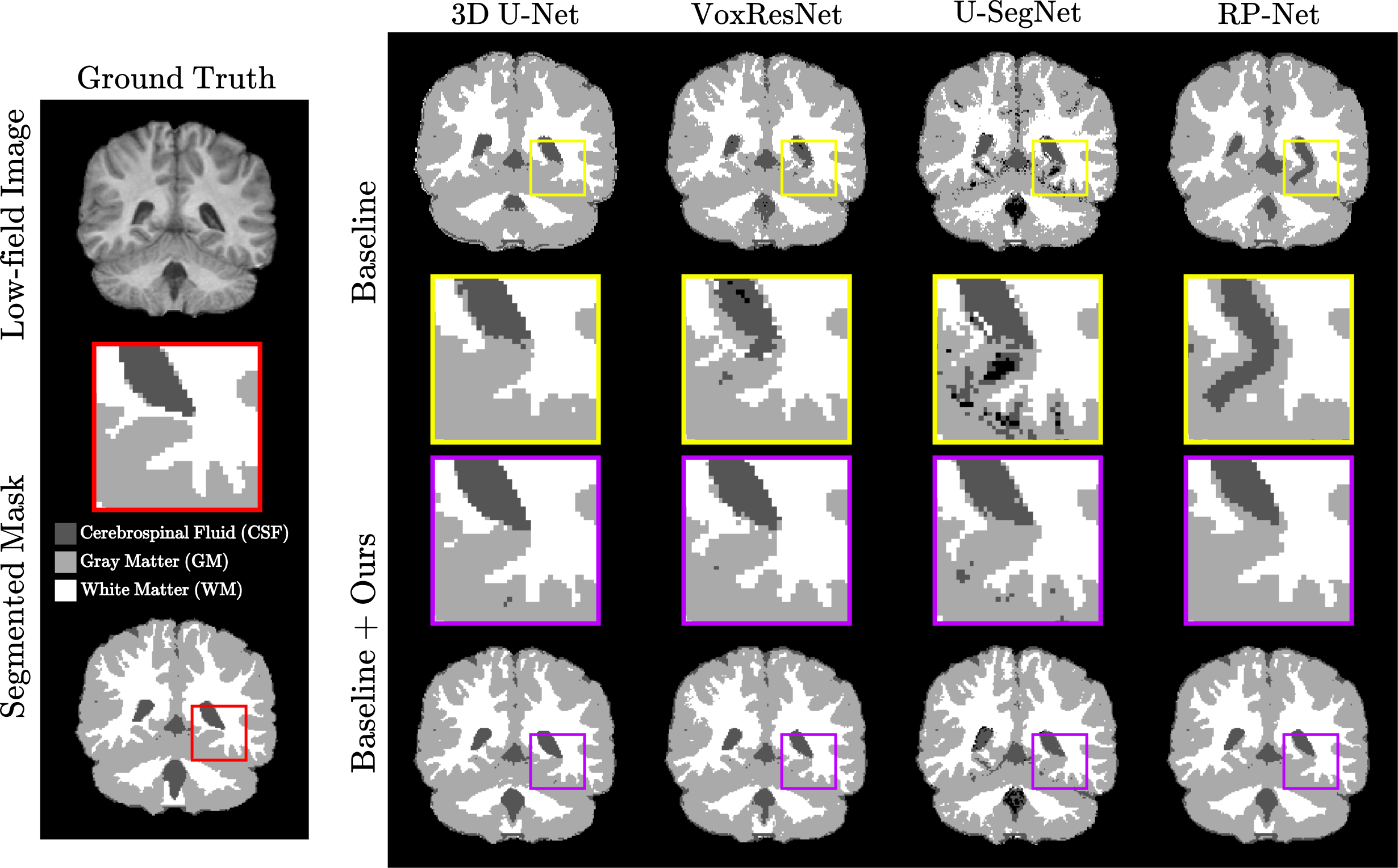}
\caption{Comparison of brain tissue segmentation results on the IBSR dataset in coronal view. The yellow boxes in the first and second rows present segmented masks predicted by baseline models, and the purple boxes in the third and fourth rows present segmented masks predicted by the proposed method applied to baseline models.}\label{qualitative_comparison_IBSR}
\end{figure}

Regarding quantitative evaluation, we measured the DSC scores from the obtained synthetic images; the results are presented in Table~\ref{tab_result_paired}. Compared to WATNet, our method using ANTs provided dramatically increased DSC scores in all tissue segments (Table~\ref{tab_result_paired}(a), (b), and (c)), and our method of learning-based segmentation decoder achieved the highest DSC performance (Table~\ref{tab_result_paired}(e)). There was an averaged DSC performance gap of approximately 2$\%$ between ANTs (Table~\ref{tab_result_paired}(b) and (c)) and U-Net decoder segmentation (Table~\ref{tab_result_paired}(d) and (e)). This outcome indicates that exploiting UHF feature representations is more beneficial than image-level information and can contribute to synthesis and segmentation tasks with the KKN. Such a visual and numerical investigation confirmed that our method achieved a clear improvement in the 7T contrast depiction, thereby providing more reliable and elaborate UHF representations.

\subsubsection{Adaptability on Brain Tissue Segmentation} 
To thoroughly assess the adaptability of UHF feature representations, we conducted a comprehensive evaluation using the proposed method on the IBSR dataset, which focuses on brain tissue segmentation. Specifically, our approach engaged in integrating the pre-trained KKN and AFMs into four existing 3D-based tissue segmentation models: 3D U-Net \citep{cciccek20163d}, VoxResNet \citep{chen2018voxresnet}, U-SegNet \citep{kumar2018u}, and RP-Net \citep{wang2019rp}. Such an integration aimed to compare the impact of applying our UHF guidance against scenarios that were not using our method (\ie, four baseline models).

\begin{table}[t!]\centering\scriptsize\setlength{\tabcolsep}{8pt}
    \caption{DSC of brain tissue segmentation on the IBSR dataset. The checkmark indicates the application of our UHF guidance to the baseline models.}\label{tab_result_IBSR}
    \begin{tabular}{cccccc}
        \toprule
         \multirow{2}{*}{\textbf{UHF}} & \multicolumn{1}{c}{\multirow{2.5}{*}{\textbf{Methods}}} & \multicolumn{4}{c}{\textbf{DSC Score ($\%$)}}\\
         \cmidrule(lr){3-6}
         \multirow{1}{*}{\textbf{Guidance}} &  & \textbf{CSF} & \textbf{GM} & \textbf{WM} & \textbf{Average}\\
        \midrule
        & 3D U-Net & 82.09 & 91.81 & 88.49 & 87.46 \\
        & VoxResNet & 77.70 & 90.72 & 86.75 & 85.06 \\
        & U-SegNet & 72.03 & 87.79 & 84.22 & 81.35 \\
        & RP-Net & 74.89 & 87.29 & 83.41 & 81.86 \\
        \midrule
        \checkmark & 3D U-Net & 84.16 (2.07$\uparrow$) & 92.52 (0.71$\uparrow$) & 88.83 (0.34$\uparrow$) & 88.50 (1.04$\uparrow$)\\
        \checkmark & VoxResNet & 81.11 (3.41$\uparrow$) & 91.75 (1.03$\uparrow$) & 87.84 (1.09$\uparrow$) & 86.90 (1.84$\uparrow$) \\
        \checkmark & U-SegNet & 76.10 (4.07$\uparrow$) & 89.41 (1.62$\uparrow$) & 85.45 (1.23$\uparrow$) & 83.65 (2.30$\uparrow$) \\
        \checkmark & RP-Net & 81.58 (6.69$\uparrow$) & 92.22 (4.93$\uparrow$) & 88.89 (5.48$\uparrow$) & 87.56 (5.70$\uparrow$) \\
        \bottomrule
    \end{tabular}
\end{table}

With the incorporation of our UHF guidance, the results of baseline models showed significant improvements in segmentation accuracy along with visual quality as illustrated in Figure~\ref{qualitative_comparison_IBSR}. For 3D U-Net, there was a notable enhancement in the clarity of edges and tissue boundaries such that it yielded more distinct and precise segmented areas. In the cases of VoxResNet and U-SegNet, we observed a marked reduction in the occurrence of noisy and inconsistent segments, leading to cleaner and more coherent segmentation outputs. Most impressively, RP-Net, which benefited significantly from our method, alleviated the mis-segmentation phenomena that caused a critical performance decrease in certain areas. To further analyze the superiority of UHF guidance, we extensively investigated the effectiveness of UHF guidance and intermediate outcomes of the lowest-level AFM as detailed in \ref{app:effectiveness-uhf-guidance-7Tabsent} and \ref{app:verification-guided-feature-ibsr-malc}. Regarding numerical assessment, Table~\ref{tab_result_IBSR} demonstrates that such a visual quality improvement also boosted the quantitative score in other baselines. In particular, RP-Net drastically enhanced the DSC scores, averaging $5.7\%$ across all tissues. From these exhaustive evaluations, we conclude that our UHF guidance not only promotes the segmentation capabilities of existing models by providing distinguishable 7T representations but also demonstrates remarkable adaptability across various segmentation models, irrespective of their underlying architecture and design.

\begin{figure}[t!]
\centering
\includegraphics[width=\textwidth]{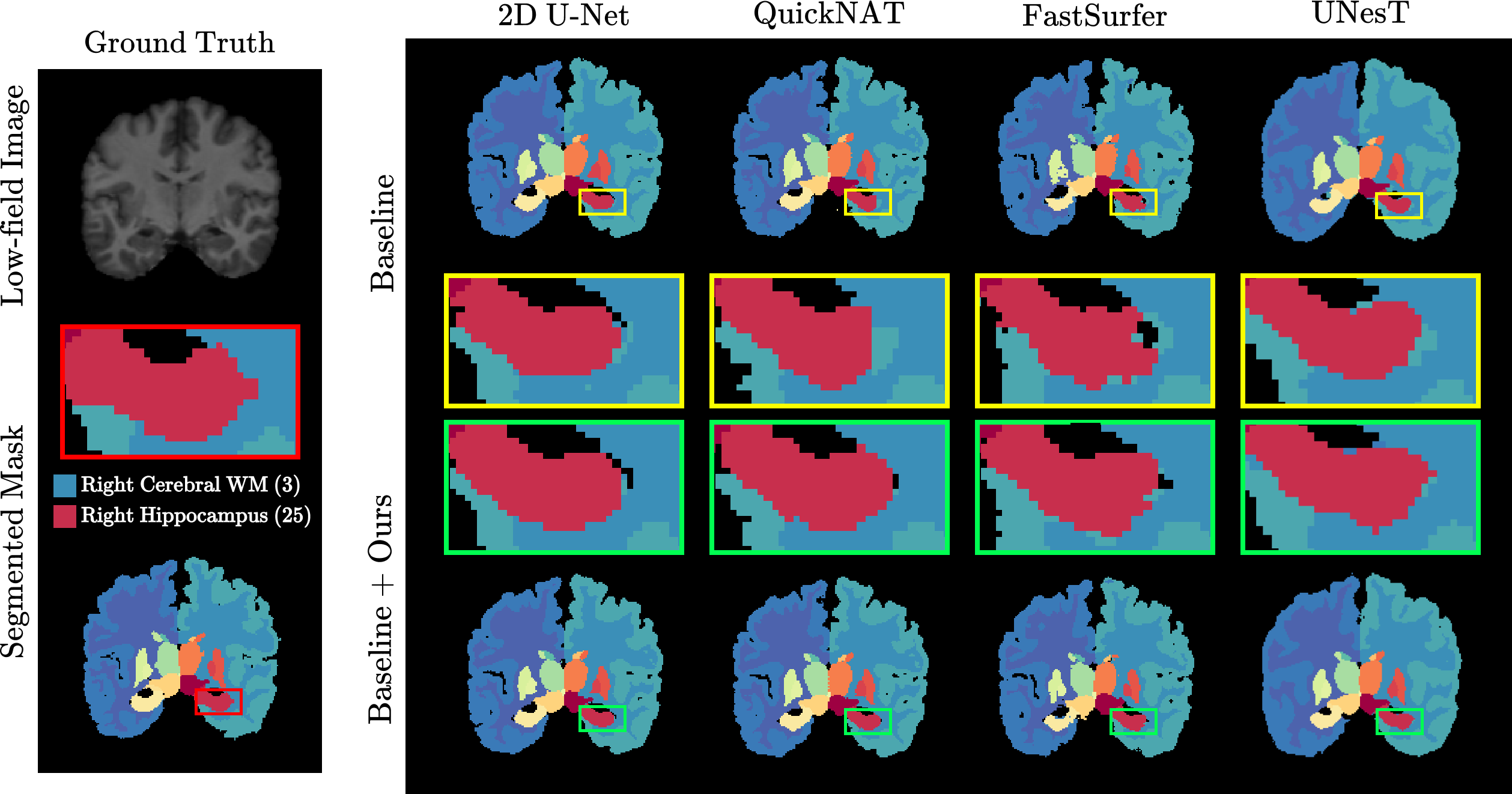}
\caption{Comparison of whole-brain segmentation results on the MALC dataset in coronal view. The yellow boxes in the first and second rows exhibit segmented masks predicted by baseline models, and the green boxes in the third and fourth rows exhibit segmented masks predicted by the proposed method applied to baselines.}\label{qualitative_comparison_MALC}
\end{figure}

\subsubsection{Scalability on Whole-Brain Segmentation} 
On the basis of the abovementioned favorable capabilities of the UHF guidance, our investigation was extended to the complex domain of whole-brain segmentation, partitioning brain regions into 27 distinct cortical and subcortical. As for validating UHF guidance in this intricate task, we employed the 2D KKN and AFMs to incorporate them into each of three comparative models: 2D U-Net \citep{ronneberger2015u}, QuickNAT \citep{roy2019quicknat}, and FastSurfer \citep{henschel2020fastsurfer}. For UNesT \citep{yu2023unest}, we leveraged a 3D KKN and AFMs depending on their compatibility according to the use of 3D MR images.

\begin{figure}[ht!]
\centering
\includegraphics[width=.775\textwidth]{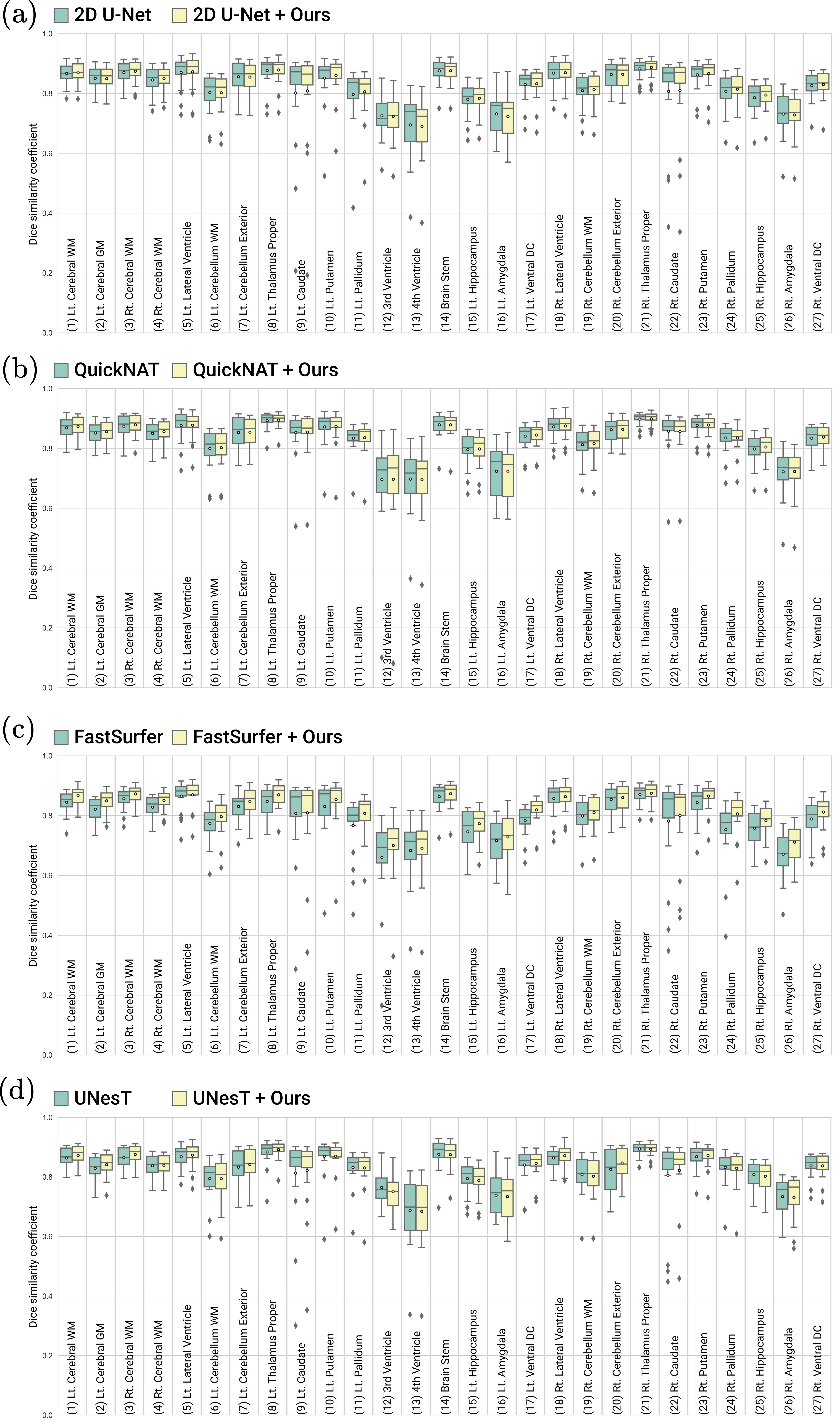}
\vspace{-0.2cm}
\caption{Illustration of box plots over the DSC scores of whole-brain segmentation.}\label{box_plot_MALC}
\end{figure}

In Figure~\ref{qualitative_comparison_MALC}, we visualized segmented masks and zoomed in on the right hippocampus, a region that exhibited notable quantitative enhancements across all baseline models with our method. From 2D U-Net and UNesT, we confirmed that applying our method improved the clarity in tissue boundaries, which was similar to the tendency in 3D U-Net results for brain tissue segmentation. Notably, QuickNAT resolved partial occlusions and truncation in the hippocampus segment, which had been an issue in its baseline segmentation mask. By effectively alleviating the noisy issue, FastSurfer yielded more clearly delineated and accurately defined hippocampal structures. For a quantitative evaluation, we have presented dice scores of the overall performance in Table~\ref{tab_result_MALC} by separating all brain segments into major structural parts of the brain: cortical (from (1) to (4) segments in Figure~\ref{box_plot_MALC}) and subcortical (from (5) to (27) segments in Figure~\ref{box_plot_MALC}). Here, we found that FastSurfer and UNesT with our method achieved an outstanding performance promotion of 2.05$\%$ and 0.72$\%$ on average, respectively, and more noticeable performance improvement was confirmed in the cortical segment rather than the densely parcellated subcortical segment overall across all baseline models.

\begin{table}[t!]\centering\scriptsize\setlength{\tabcolsep}{10pt}
    \caption{Averaged DSC of cortical and subcortical structures of whole-brain segmentation on the MALC dataset. Here, the checkmark indicates the application of our UHF guidance to the baseline models. ``Cortical'' refers to four regions, including cerebral WM and GM on both the left and right hemispheres of the brain, while ``Subcortical'' refers to the remaining 23 regions.}\label{tab_result_MALC}
    \begin{tabular}{cccccc}
        \toprule
        \multirow{2}{*}{\textbf{UHF}} & \multicolumn{1}{c}{\multirow{2.5}{*}{\textbf{Methods}}} & \multicolumn{3}{c}{\textbf{DSC Score ($\%$)}}\\
        \cmidrule(lr){3-5}
        \multirow{1}{*}{\textbf{Guidance}} & & \textbf{Cortical} & \textbf{Subcortical} & \textbf{Average}\\
        \midrule
        & 2D U-Net & 85.80 & 81.50 & 82.14 \\
        & QuickNAT & 86.11 & 82.54 & 83.08 \\
        & FastSurfer & 83.75 & 78.96 & 79.67 \\
        & UNesT & 86.86 & 82.07 & 82.49 \\
        \midrule
        \checkmark & 2D U-Net & 86.07 (0.27$\uparrow$) & 81.68 (0.18$\uparrow$) & 82.33 (0.19$\uparrow$)\\
        \checkmark & QuickNAT & 86.57 (0.46$\uparrow$) & 82.66 (0.12$\uparrow$) & 83.24 (0.16$\uparrow$) \\
        \checkmark & FastSurfer & 85.98 (2.23$\uparrow$) & 80.98 (2.02$\uparrow$) & 81.72 (2.05$\uparrow$) \\
        \checkmark & UNesT & 85.71 (1.15$\uparrow$) & 82.49 (0.42$\uparrow$) & 83.21 (0.72$\uparrow$) \\
        \bottomrule
    \end{tabular}
\end{table}

We further analyzed the variance of segmentation performance according to the region-wise DSC scores as shown in Figure~\ref{box_plot_MALC}. In both U-Net and QuickNAT, we found a general trend of performance increase across most segments. However, there was a slight performance decline in smaller brain regions, notably in the (13) 4th ventricle for both models. This observation suggests that our AFM might have struggled to provide the UHF guidance effectively because the ventricles of the brain represented not only a very dense area but also a colorless region filled with CSF. Meanwhile, FastSurfer and UNesT with our method revealed a marked boost in DSC scores across all brain regions relative to its baseline performance. Consequently, these qualitative and quantitative outcomes comprehensively validate the scalability of the proposed method, highlighting its capacity to be applied beyond simple tissue segmentation to detailed brain region segmentation tasks. The findings could provide compelling evidence that the consistency of performance gains, especially in more intricate brain regions, supports the robustness and versatility of our approach.


\subsection{Ablation Study} 
To explore the effectiveness of our method, we designed several scenarios by ablating each principal component within the KKN and AFM. For the investigation of the KKN components, we first trained the variant network for a task of 7T-like image synthesis on the paired 3T-7T dataset. Afterward, the pre-trained network was employed for brain tissue segmentation, utilizing RP-Net \citep{wang2019rp}: a model that had demonstrated substantial improvements on the IBSR dataset compared to other baselines. Similarly, the vital components of AFM were also analyzed by brain tissue segmentation to verify the efficient assimilation of UHF guidance into LF features.

\begin{table}[t!]\centering\scriptsize\setlength{\tabcolsep}{3.5pt}
    \caption{Ablation on each component of the KKN in 7T-like image synthesis on the paired 3T-7T dataset and in brain tissue segmentation on the IBSR dataset. In the segmentation results, the DSC is used for assessment, and RP-Net is adopted as the baseline model.}\label{tab_ablation_KKN}
    \begin{tabular}{lccccccc}
        \toprule
        \multicolumn{1}{c}{\multirow{2.7}{*}{\textbf{Ablation Components}}} & \multicolumn{2}{c}{\textbf{Image Synthesis}} & \multicolumn{4}{c}{\textbf{Segmentation ($\%$)}}\\
        \cmidrule(lr){2-3} \cmidrule(lr){4-7}
        & \textbf{PSNR} & \textbf{SSIM} & \textbf{CSF} & \textbf{GM} & \textbf{WM} & \textbf{Average}\\
        \midrule
        (a) w/o Guide Blocks & 26.03 & 0.9239 & 81.24 & 92.05 & 88.37 & 87.22\\
        (b) w/o Knowledge Distillation & 26.15 & 0.9265 & 81.44 & 91.79 & 87.89 & 87.04\\
        (c) Ours & \textbf{26.43} & \textbf{0.9266} & \textbf{81.58} & \textbf{92.22} & \textbf{88.89} & \textbf{87.56}\\
        \bottomrule
    \end{tabular}
\end{table}

\subsubsection{Ablating Principal Components in KKN}
We conducted an in-depth quantitative analysis to assess the impact of ablating guide blocks or KD from the KKN. This evaluation was crucial for understanding the network architecture and training strategy strengths of the KKN. The results in Table~\ref{tab_ablation_KKN} reveal a performance degradation in both 7T-like image synthesis (in the presence of 7T data) and segmentation tasks (in the absence of 7T data) when these components were excluded.

Specifically, the ablation of guide blocks (Table~\ref{tab_ablation_KKN}(a)), which play a pivotal role in recalibrating concatenated 3T features, slightly reduced the capability of the KKN to synthesize 7T-like images and performed accurate segmentation effectively. This result suggests that the guide blocks influence the optimal alignment between the 3T features and the UHF feature representations. Moreover, eliminating the KD strategy (Table~\ref{tab_ablation_KKN}(b)), which allows indirect learning from 7T image features via the teacher network, further exacerbated the performance drop. This finding underlines the importance of KD in facilitating the transfer of intricate UHF feature representations from the teacher to the student network in the 7T-absent domain. One interesting observation in KD ablation was the inverse relationship between image synthesis and segmentation performance. Methods excelling in image synthesis did not necessarily translate to superior segmentation outcomes, particularly for GM and WM. Therefore, we assume that image synthesis requires high fidelity in reproducing visual details, whereas segmentation necessitates a nuanced comprehension of anatomical structures, which might not directly correlate with visual authenticity. Integrating these two components (Ours in Table~\ref{tab_ablation_KKN}(c)) afforded the best performance across both tasks. Such success can be attributed to the capabilities of the KKN: learning UHF feature representations in a 7T-present domain (image synthesis) and effectively transforming LF images into 7T-like features in a 7T-absent domain (segmentation).

\begin{table}[t!]\centering\scriptsize\setlength{\tabcolsep}{3.5pt}
    \caption{Ablation on each component of the AFM in brain tissue segmentation on the IBSR dataset. We used DSC as the evaluation matrix and employed RP-Net as the baseline model.}\label{tab_ablation_fusion}
    \begin{tabular}{lccccc}
        \toprule
        \multicolumn{1}{c}{\multirow{2.7}{*}{\textbf{Ablation Components}}} & \multicolumn{4}{c}{\textbf{DSC Score ($\%$)}}\\
        \cmidrule(lr){2-5}
         & \textbf{CSF} & \textbf{GM} & \textbf{WM} & \textbf{Average}\\
        \midrule
        (a) w/o Aggregation & 81.24 & {91.98} & {88.36} & 87.19\\
        (b) w/o Two Paths in Channel-wise Refinement & 81.72 & 91.89 & 88.35 & 87.32\\
        (c) w/o Feature-combined Refinement & {82.25} & 91.79 & 88.11 & 87.38\\
        (d) w/o Channel-wise Refinement & \textbf{82.49} & 91.92 & 87.93 & {87.45}\\
        (e) Ours & 81.58 & \textbf{92.22} & \textbf{88.89} & \textbf{87.56}\\
        \bottomrule
    \end{tabular}
\end{table}

\subsubsection{Ablating Principal Components in AFM} 
We conducted a series of ablation experiments to determine the role and influence of each component within the AFM on generating assimilable UHF guidance for LF features (Table~\ref{tab_ablation_fusion}). To this end, we designed the four ablation scenarios depending on the core components of AFM: knowledge aggregation, feature-combined refinement, and channel-wise refinement, including two ways that partition the guidance according to the sign of values.

One of the most prominent findings from these experiments was the considerable influence of knowledge aggregation. The ablation of this component yielded a markedly reduced performance, which was particularly noticeable in the diminished DSC scores for the CSF segment (Table~\ref{tab_ablation_fusion}(a)). As this component reflects both the fine and coarse details necessary for precise brain segmentation, we speculated that an ablating aggregation component could cause performance degradation owing to a lack of comprehensive feature utilization at different hierarchical levels of 7T-like representations. Meanwhile, we observed the trade-off between CSF and WM segmentation in several scenarios (Table~\ref{tab_ablation_fusion}(b), (c), and (d)). This imbalance issue suggests that the guided features tend to be biased toward either the brighter nerve fibers (WM) or the more colorless fluid-filled spaces (CSF) unless the feature refinement strategy is properly employed in our AFM. The CSF particularly relies heavily on contrast differences because this region is less structurally complex than other brain tissues. Accordingly, harmonizing such refinement phases allows the AFM to be more effective in enhancing the segmentation accuracy from these tissue types by alleviating the bias phenomena. In this context, the proposed AFM obtained appropriately guided features with a balanced improvement of all tissue segments, resulting in the best performance on average. This demonstrates that the AFM can effectively adjust the contrast and structural details via UHF guidance while preserving the inherent characteristics of LF features.

\section{Discussion and Conclusion}
\label{section:conc}
In this study, we proposed a novel adaptive-fusion strategy to adequately adjust the contrast according to intensity values of LF features by UHF guidance so that arbitrary segmentation models could better uncover intricately detailed structures that are typically arduous to perceive when relying solely on LF features. In particular, our strategy has impressively explored the elaborate balance between maintaining inherent image structures and managing excessive contrast in the 7T-absent domain. In detail, we have mitigated this issue by designing a knowledge flow that aggregates and assimilates UHF feature representations, imposing specialized UHF guidance tailored to the specific needs of given LF features. In the quantitative and qualitative evaluations, we showed that the proposed method outperformed all baseline models along with the superior visual quality of segmentation masks. Moreover, our AFM with a pre-trained KKN not only demonstrated adaptability in brain tissue segmentation but also exhibited remarkable scalability in whole-brain segmentation. Such beneficial capabilities were obvious in how our method successfully integrated with various baseline models, enhancing their ability to discern fine details in brain tissue structures often disregarded in LF images while preserving structural LF appearance.

The Segment Anything Model (SAM)~\citep{kirillov2023segment}, a recently introduced foundation model, has garnered attention for its promising zero-shot segmentation capabilities. Notably, this model is able to segment not only natural image datasets but also extend its proficiency to medical imaging~\citep{mazurowski2023segment,huang2024segment}. However, despite such impressive superiority, SAM comes with its own set of limitations. It operates on a 2D basis, which means that 3D segmentation tasks necessitate processing images slice by slice and then reconstructing them in 3D form. This procedure of transitioning from 2D to 3D often carries the risk of introducing errors, which could potentially compromise the accuracy and reliability of the resultant 3D segmentation. In addition, SAM still has room for improvement in brain segmentation tasks that differentiate intricate whole-brain regions~\citep{huang2024segment}. By contrast, the proposed method exhibits exceptional versatility, functioning effectively across various data sizes (\ie, regardless of data dimensionality) while enabling the complex task of whole-brain segmentation.


Although our current focus is confined to structural brain image segmentation by utilizing UHF feature representations, it is important to note the broader facilitates of UHF imaging. In this regard, UHF imaging is renowned for its potential to reveal lesions that are undetectable in LF images \citep{nielsen2013contribution,shaffer2022ultra}. This ability allows our method to be extensively applied to brain tumor segmentation such that it can help detect lesions by employing UHF feature representations even learned from healthy adult brains imaged at 7T. However, we anticipate that our method could also be adapted for MRI super-resolution applications, with the aim to reconstruct high-resolution representations from low-resolution scans affected by subject motion or scanning time constraints \citep{van2012super}. Such an intuitive understanding of the impact and limitations of our UHF guidance approach enables us to provide valuable insights for future enhancements and applications in brain image segmentation. In conclusion, we believe that this endeavor and expansion holds promise in elevating the scope for quality and precision of neuroimaging, enriching the field of medical imaging with more exhaustive and detailed analyses.

\section*{Code and Data Availability}
Our implementation source code is available at \url{https://github.com/ku-milab/UHF-guided_segmentation}. Data used in the brain tissue and whole-brain segmentation were respectively obtained from the Internet Brain Segmentation Repository at \url{https://www.nitrc.org/projects/ibsr} and Multi-Atlas Labeling Challenge at \url{https://neuromorphometrics.com/}. Contrarily, the paired 3T-7T dataset has not been published. The raw sMRI data are protected and are not made public owing to data privacy laws, yet they are available from the corresponding author upon reasonable request.

\section*{Declaration of Competing Interest}
The authors declare that they have no conflict of interest.

\section*{CRediT Authorship Contribution Statement}
\textbf{Kwanseok Oh:} Methodology, Software, Validation, Formal analysis, Investigation, Writing - original draft preparation, Project administration. \textbf{Jieun Lee:} Conceptualization, Methodology, Software, Validation, Formal analysis, Investigation, Writing - original draft preparation, Visualization. \textbf{Da-Woon Heo:} Software, Validation, Formal analysis, Writing - review \& editing. \textbf{Dinggang Shen:} Data Curation, Writing - review \& editing. \textbf{Heung-Il Suk:} Conceptualization, Methodology, Validation, Writing - review \& editing, Supervision, Project administration, Funding acquisition.

\section*{Acknowledgements}
This work was supported by the Institute of Information \& communications Technology Planning \& Evaluation (IITP) grant funded by the Korea government (MSIT) No. 2022-0-00959 ((Part 2) Few-Shot Learning of Causal Inference in Vision and Language for Decision Making) and No. 2019-0-00079 (Department of Artificial Intelligence (Korea University)).

\clearpage
\appendix

\section{Verification of 7T-like Feature Representations on Paired 3T-7T Dataset}\label{app:verification-7t-like-feature}
We investigated the differences and similarities in intermediate feature maps extracted by the KKN and teacher network to understand UHF feature representations better. Specifically, our KKN employs two core modules: a feature extractor for capturing LF features from 3T images and a guide block dedicated to transforming these LF features into UHF feature representations. We can thus categorize the resultant feature maps into 3T features (derived from the LF feature extractor) and 7T-like features (produced by the guide block). In addition, pure 7T features could be directly obtained from the teacher network, which processes 7T images for comparison with our 7T-like features.

To compare three features (\ie, 3T, 7T, and 7T-like features) across all levels from coarse to fine, we averaged the respective feature maps obtained from the 3D KKN or teacher, followed by calculating the L2 distance at each hierarchical level. According to the results presented in Table~\ref{app:tab_fmap_paired}(a), the 3T and 7T features exhibited the most significant divergence across all levels, reflecting the inherent disparity between LF and UHF imaging. Interestingly, although the 7T-like features originated from 3T images, they exhibited in a considerable departure from the 3T features (Table~\ref{app:tab_fmap_paired}(b)) and a much closer resemblance to 7T features (Table~\ref{app:tab_fmap_paired}(c)). These observations suggest that 3T images were successfully converted into UHF-level feature representations similar to those derived from 7T images.

Further visualization of these transformed feature maps is provided in Figure~\ref{app:featuremap_visualization}, and the 7T-like features generated by the KKN are particularly noteworthy in terms of visual quality. Regions that initially exhibited weak intensity in both the 3T images and their corresponding 3T features underwent a significant enhancement, with a contrast level suggestive of 7T features. This transformation highlights the efficacy of the KKN in converting LF features into a UHF feature representation.

\setcounter{table}{0}
\begin{table}[h]\centering\scriptsize\setlength{\tabcolsep}{5.5pt}
    \caption{Euclidean distance (\ie, L2-norm) of each level between (a) 3T and 7T features, (b) 3T and 7T-like features, and (c) 7T and 7T-like features. The results are reported across 15 folds (mean$\pm$standard deviation).}\label{app:tab_fmap_paired}
    \begin{tabular}{cccccc}
        \toprule
        & \textbf{Level 1} & \textbf{Level 2} & \textbf{Level 3} & \textbf{Level 4} & \textbf{Level 5}\\
        \midrule
        (a) & 0.0617±0.0608 & 0.0351±0.0186 & 0.0241±0.0082 & 0.0390±0.0183 & 0.0500±0.0296\\
        (b) & 0.0583±0.0810 & 0.0443±0.0393 & 0.0181±0.0074 & 0.0297±0.0138 & 0.0410±0.0201\\
        (c) & \textbf{0.0183}±\textbf{0.0105} & \textbf{0.0165}±\textbf{0.0147} & \textbf{0.0083}±\textbf{0.0055} & \textbf{0.0076}±\textbf{0.0053} & \textbf{0.0200}±\textbf{0.0092}\\
        \bottomrule
    \end{tabular}
\end{table}

\setcounter{figure}{0}
\begin{figure}[h]
\centering
\includegraphics[width=0.8\textwidth]{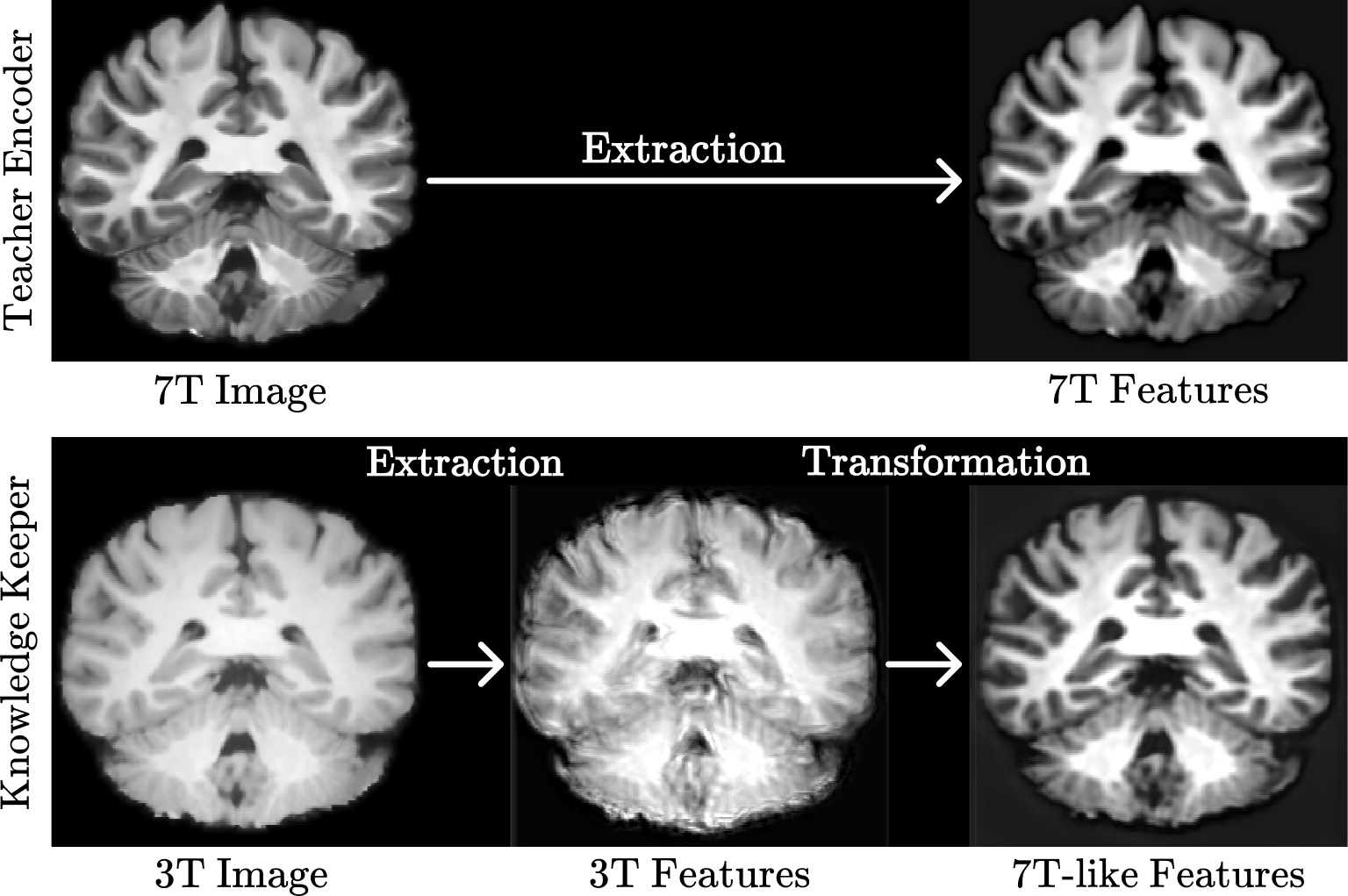}
\caption{Visualization of the averaged feature maps extracted or transformed at the lowest level (Level 1) by the teacher encoder or the knowledge keeper (\ie, our KKN).}\label{app:featuremap_visualization}
\end{figure}

\clearpage
\section{Effectiveness of Feature-level UHF Guidance in 7T-Absent Domain}\label{app:effectiveness-uhf-guidance-7Tabsent}
To thoroughly demonstrate the effectiveness of UHF guidance in the 7T-absent domain, we designed a comprehensive evaluation procedure separated into three scenarios: (a) image-level UHF guidance, (b) baseline (\ie, without UHF guidance), and (c) feature-level UHF guidance. For the image-level approach, we synthesized the input LF image using the KKN and teacher decoder. Subsequently, this synthesized image was used to train 3D U-Net, one of the baseline segmentation models; the predicted segmentation mask derived from the trained 3D U-Net is depicted at the top of Figure~\ref{comparison_image-feature_IBSR}. For a comparison, the output of the feature-level approach is visualized at the bottom of Figure~\ref{comparison_image-feature_IBSR}.

As revealed by Table~\ref{tab_result_imglevel_IBSR}, such an image-level approach yielded inferior quantitative results, approximately 6.8$\%$ lower than the baseline model trained without UHF guidance, which is significant. We hypothesized that the poor performance of the image-level approach could largely be attributed to the distortion or alteration of inherent LF features during the image-to-image translation process. This effect is visually substantiated in Figure~\ref{comparison_image-feature_IBSR}, where the segmented masks produced using the image-level approach demonstrate blurred tissue edges, indicating the loss of structural integrity in the LF image and the exaggerated contrast in the synthesized image. Contrary to this approach, our feature-level UHF guidance effectively captures the nuanced representation of the brain's anatomical structures while reflecting the intrinsic characteristics of LF images without distortion. Moreover, the DSC score at the feature-level UHF guidance was on average 7.84$\%$ higher than that with the image-level approach. On the basis of these exhaustive analyses, we believe that applying UHF guidance at the feature level is a more strategic and practical approach. By integrating UHF information directly with inherent LF features, this strategy preserves essential structural elements of LF images while leveraging the detailed and high-quality characteristics of UHF imaging to ensure more accurate and reliable segmentation.

\setcounter{table}{0}
\begin{table}[h]\centering\scriptsize\setlength{\tabcolsep}{13pt}
    \caption{Dice similarity coefficient (DSC) of brain tissue segmentation of 3D U-Net when using UHF guidance at the image level or feature level on the IBSR dataset.}\label{tab_result_imglevel_IBSR}
    \begin{tabular}{cccccc}
        \toprule
        & \multirow{2.7}{*}{\textbf{Use of UHF Guidance}} & \multicolumn{4}{c}{\textbf{DSC Score ($\%$)}}\\
         \cmidrule(lr){3-6}
        & & \textbf{CSF} & \textbf{GM} & \textbf{WM} & \textbf{Average}\\
        \midrule
        (a) & Image level & 70.46 & 87.79 & 83.74 & 80.66\\
        (b) & None (Baseline) & 82.09 & 91.81 & 88.49 & 87.46\\
        (c) & Feature level & \textbf{84.16} & \textbf{92.52} & \textbf{88.83} & \textbf{88.50}\\
        \bottomrule
    \end{tabular}
\end{table}

\setcounter{figure}{0}
\begin{figure}[h]
\centering
\includegraphics[width=1.\textwidth]{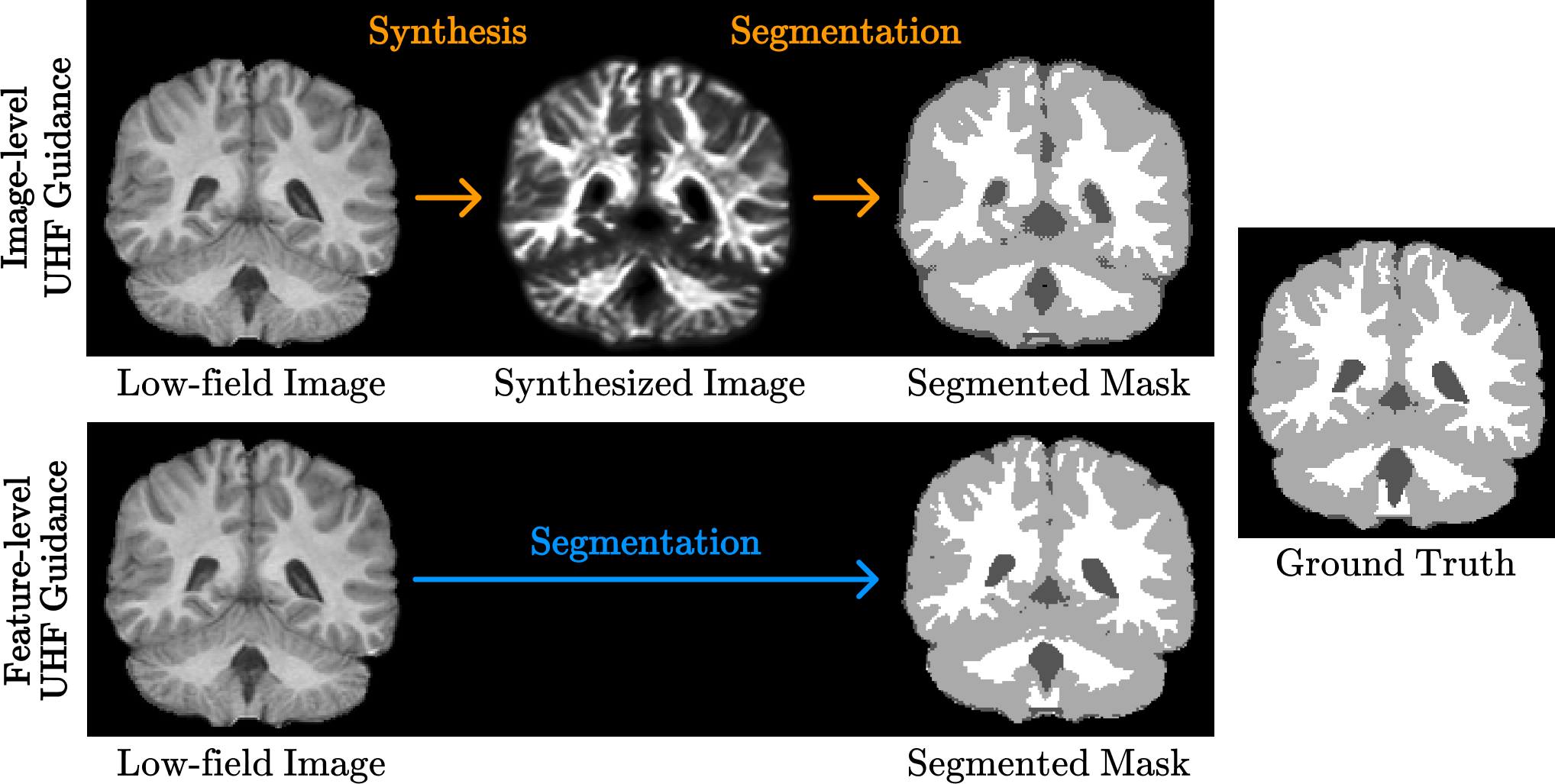}
\caption{Comparison of brain tissue segmentation results of 3D U-Net between the image-level and feature-level UHF guidance on the IBSR dataset.}\label{comparison_image-feature_IBSR}
\end{figure}

\clearpage
\section{Verification of Guided Feature Representations on IBSR and MALC datasets}\label{app:verification-guided-feature-ibsr-malc}
To gain an intuitive understanding of the direct impact of UHF guidance, we thoroughly analyzed the intermediate feature maps of our AFM (\ie, feature maps that applied the UHF guidance) on IBSR and MALC datasets. To this end, we selected the outcomes of the lowest-level AFM and visualized them as shown in Figures~\ref{guided_features_IBSR} and \ref{guided_features_MALC}. The first and third rows in each figure present the LF features before and after applying the guided features in the second row, respectively. Across all baseline models in these qualitative assessments, we observed that UHF guidance tended to selectively enhance or diminish the intensity in certain regions while retaining other inherent characteristics, such as brain structure, which were well-represented in LF features. From these investigations, we confirmed that segmentation performance could be enhanced by the capabilities of UHF-guided features that delineate and highlight specific anatomical features, resulting in more accurate and detailed segmentation results.

\setcounter{figure}{0}
\begin{figure}[h]
\centering
\includegraphics[width=0.85\textwidth]{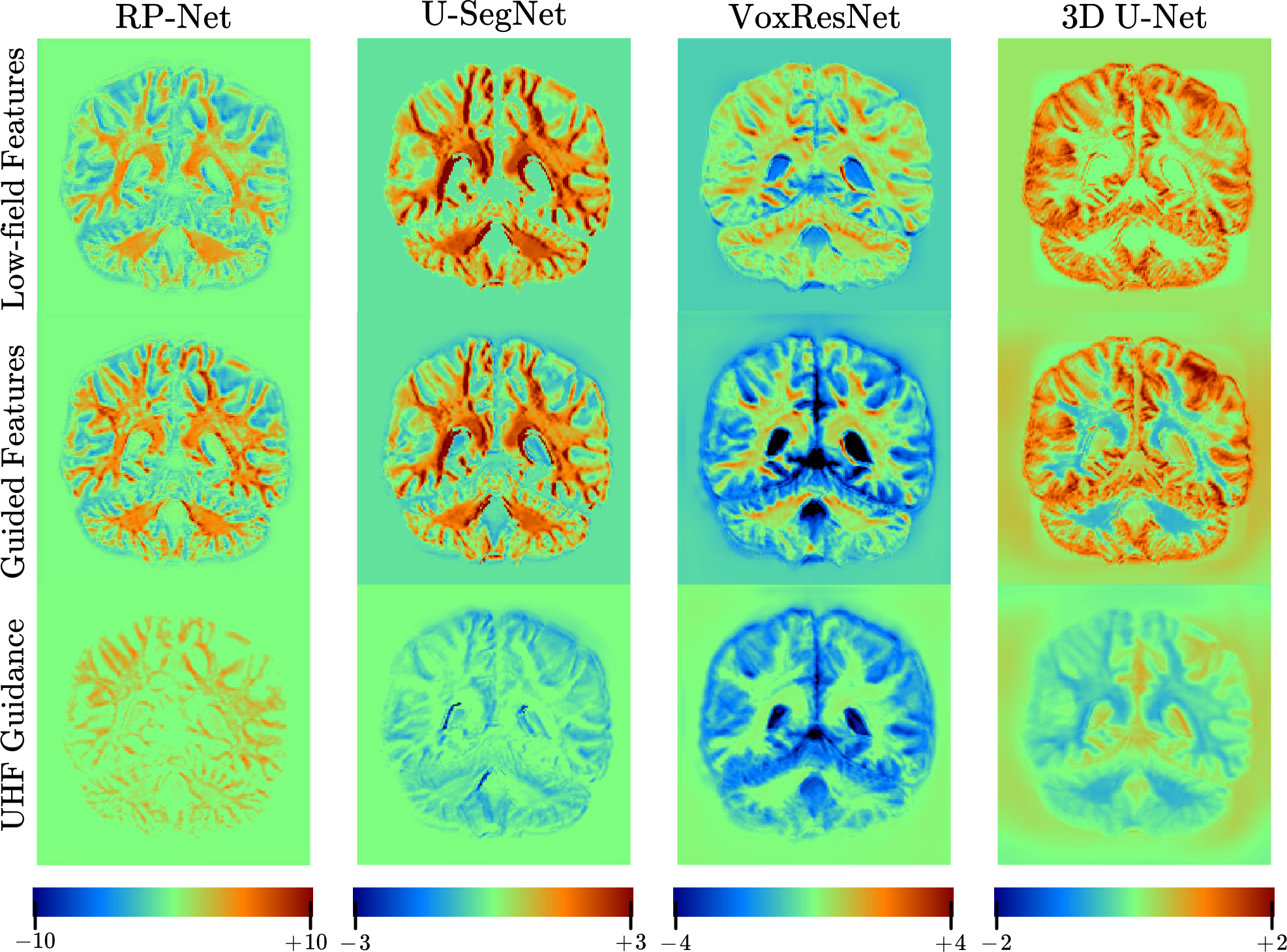}
\caption{Visualization of the lowest-level feature maps from our AFM trained on the IBSR dataset. In AFM, guided features are obtained by adding low-field features and UHF guidance. Each column refers to a different baseline model that applied our method.}\label{guided_features_IBSR}
\end{figure}

\setcounter{figure}{1}
\begin{figure}[h]
\centering
\includegraphics[width=0.85\textwidth]{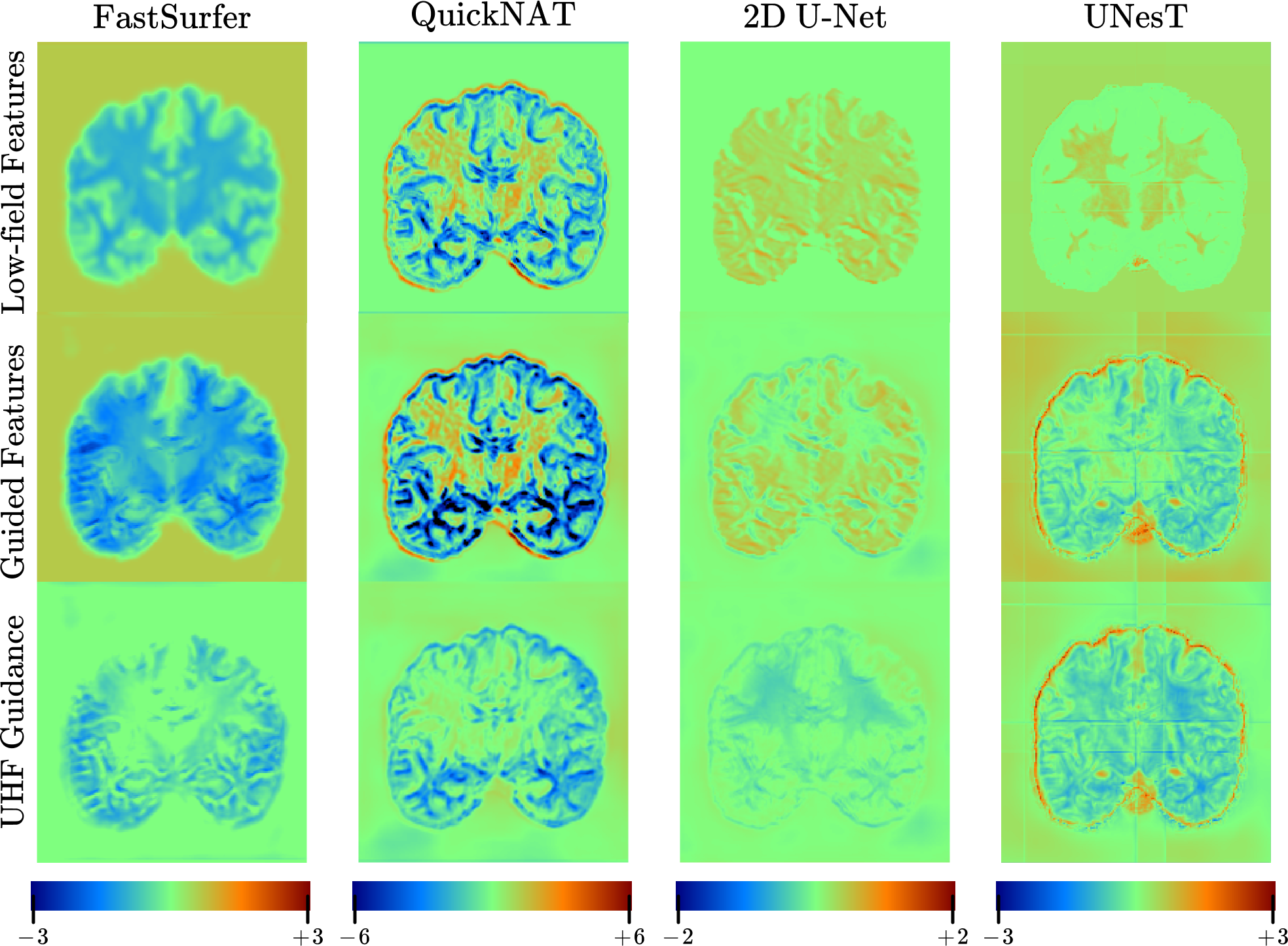}
\caption{Visualization of the lowest level's feature maps from our AFM trained on the MALC dataset. In AFM, guided features are obtained by adding low-field features and UHF guidance. Each column refers to a different baseline model that applied our method.}\label{guided_features_MALC}
\end{figure}

\clearpage
\bibliographystyle{elsarticle-harv} 
\bibliography{main}

\begin{thebibliography}{45}
\expandafter\ifx\csname natexlab\endcsname\relax\def\natexlab#1{#1}\fi
\providecommand{\url}[1]{\texttt{#1}}
\providecommand{\href}[2]{#2}
\providecommand{\path}[1]{#1}
\providecommand{\DOIprefix}{doi:}
\providecommand{\ArXivprefix}{arXiv:}
\providecommand{\URLprefix}{URL: }
\providecommand{\Pubmedprefix}{pmid:}
\providecommand{\doi}[1]{\href{http://dx.doi.org/#1}{\path{#1}}}
\providecommand{\Pubmed}[1]{\href{pmid:#1}{\path{#1}}}
\providecommand{\bibinfo}[2]{#2}
\ifx\xfnm\relax \def\xfnm[#1]{\unskip,\space#1}\fi
\bibitem[{Avants et~al.(2011)Avants, Tustison, Wu, Cook and
  Gee}]{avants2011open}
\bibinfo{author}{Avants, B.B.}, \bibinfo{author}{Tustison, N.J.},
  \bibinfo{author}{Wu, J.}, \bibinfo{author}{Cook, P.A.}, \bibinfo{author}{Gee,
  J.C.}, \bibinfo{year}{2011}.
\newblock \bibinfo{title}{An open source multivariate framework for n-tissue
  segmentation with evaluation on public data}.
\newblock \bibinfo{journal}{Neuroinformatics} \bibinfo{volume}{9},
  \bibinfo{pages}{381--400}.
\bibitem[{Badrinarayanan et~al.(2017)Badrinarayanan, Kendall and
  Cipolla}]{badrinarayanan2017segnet}
\bibinfo{author}{Badrinarayanan, V.}, \bibinfo{author}{Kendall, A.},
  \bibinfo{author}{Cipolla, R.}, \bibinfo{year}{2017}.
\newblock \bibinfo{title}{Segnet: a deep convolutional encoder-decoder
  architecture for image segmentation}.
\newblock \bibinfo{journal}{IEEE Transactions on Pattern Analysis and Machine
  Intelligence} \bibinfo{volume}{39}, \bibinfo{pages}{2481--2495}.
\bibitem[{Bahrami et~al.(2017)Bahrami, Shi, Rekik, Gao and
  Shen}]{bahrami20177t}
\bibinfo{author}{Bahrami, K.}, \bibinfo{author}{Shi, F.},
  \bibinfo{author}{Rekik, I.}, \bibinfo{author}{Gao, Y.},
  \bibinfo{author}{Shen, D.}, \bibinfo{year}{2017}.
\newblock \bibinfo{title}{7{T}-guided super-resolution of 3{T MRI}}.
\newblock \bibinfo{journal}{Medical Physics} \bibinfo{volume}{44},
  \bibinfo{pages}{1661--1677}.
\bibitem[{Bazin et~al.(2014)Bazin, Weiss, Dinse, Sch{\"a}fer, Trampel and
  Turner}]{bazin2014computational}
\bibinfo{author}{Bazin, P.L.}, \bibinfo{author}{Weiss, M.},
  \bibinfo{author}{Dinse, J.}, \bibinfo{author}{Sch{\"a}fer, A.},
  \bibinfo{author}{Trampel, R.}, \bibinfo{author}{Turner, R.},
  \bibinfo{year}{2014}.
\newblock \bibinfo{title}{A computational framework for ultra-high resolution
  cortical segmentation at 7 {T}esla}.
\newblock \bibinfo{journal}{NeuroImage} \bibinfo{volume}{93},
  \bibinfo{pages}{201--209}.
\bibitem[{Buciluǎ et~al.(2006)Buciluǎ, Caruana and
  Niculescu-Mizil}]{bucilua2006model}
\bibinfo{author}{Buciluǎ, C.}, \bibinfo{author}{Caruana, R.},
  \bibinfo{author}{Niculescu-Mizil, A.}, \bibinfo{year}{2006}.
\newblock \bibinfo{title}{Model compression}, in:
  \bibinfo{booktitle}{Proceedings of the 12th ACM SIGKDD International
  Conference on Knowledge Discovery and Data Mining}, pp.
  \bibinfo{pages}{535--541}.
\bibitem[{Chen et~al.(2018)Chen, Dou, Yu, Qin and Heng}]{chen2018voxresnet}
\bibinfo{author}{Chen, H.}, \bibinfo{author}{Dou, Q.}, \bibinfo{author}{Yu,
  L.}, \bibinfo{author}{Qin, J.}, \bibinfo{author}{Heng, P.A.},
  \bibinfo{year}{2018}.
\newblock \bibinfo{title}{Vox{R}es{N}et: deep voxelwise residual networks for
  brain segmentation from 3{D MR} images}.
\newblock \bibinfo{journal}{NeuroImage} \bibinfo{volume}{170},
  \bibinfo{pages}{446--455}.
\bibitem[{Chen et~al.(2021)Chen, Liu, Zhao and Jia}]{chen2021distilling}
\bibinfo{author}{Chen, P.}, \bibinfo{author}{Liu, S.}, \bibinfo{author}{Zhao,
  H.}, \bibinfo{author}{Jia, J.}, \bibinfo{year}{2021}.
\newblock \bibinfo{title}{Distilling knowledge via knowledge review}, in:
  \bibinfo{booktitle}{Proceedings of the IEEE/CVF Conference on Computer Vision
  and Pattern Recognition}, pp. \bibinfo{pages}{5008--5017}.
\bibitem[{{\c{C}}i{\c{c}}ek et~al.(2016){\c{C}}i{\c{c}}ek, Abdulkadir,
  Lienkamp, Brox and Ronneberger}]{cciccek20163d}
\bibinfo{author}{{\c{C}}i{\c{c}}ek, {\"O}.}, \bibinfo{author}{Abdulkadir, A.},
  \bibinfo{author}{Lienkamp, S.S.}, \bibinfo{author}{Brox, T.},
  \bibinfo{author}{Ronneberger, O.}, \bibinfo{year}{2016}.
\newblock \bibinfo{title}{3{D U}-{N}et: learning dense volumetric segmentation
  from sparse annotation}, in: \bibinfo{booktitle}{International Conference on
  Medical Image Computing and Computer-Assisted Intervention},
  \bibinfo{organization}{Springer}. pp. \bibinfo{pages}{424--432}.
\bibitem[{Clarke et~al.(1995)Clarke, Velthuizen, Camacho, Heine, Vaidyanathan,
  Hall, Thatcher and Silbiger}]{clarke1995mri}
\bibinfo{author}{Clarke, L.}, \bibinfo{author}{Velthuizen, R.},
  \bibinfo{author}{Camacho, M.}, \bibinfo{author}{Heine, J.},
  \bibinfo{author}{Vaidyanathan, M.}, \bibinfo{author}{Hall, L.},
  \bibinfo{author}{Thatcher, R.}, \bibinfo{author}{Silbiger, M.},
  \bibinfo{year}{1995}.
\newblock \bibinfo{title}{{MRI} segmentation: Methods and applications}.
\newblock \bibinfo{journal}{Magnetic Resonance Imaging} \bibinfo{volume}{13},
  \bibinfo{pages}{343--368}.
\bibitem[{Deng et~al.(2016)Deng, Yu, Wang, Shi, Yap, Shen and
  Initiative}]{deng2016learning}
\bibinfo{author}{Deng, M.}, \bibinfo{author}{Yu, R.}, \bibinfo{author}{Wang,
  L.}, \bibinfo{author}{Shi, F.}, \bibinfo{author}{Yap, P.T.},
  \bibinfo{author}{Shen, D.}, \bibinfo{author}{Initiative, A.D.N.},
  \bibinfo{year}{2016}.
\newblock \bibinfo{title}{Learning-based 3{T} brain {MRI} segmentation with
  guidance from 7{T MRI} labeling}.
\newblock \bibinfo{journal}{Medical Physics} \bibinfo{volume}{43},
  \bibinfo{pages}{6588--6597}.
\bibitem[{Duan et~al.(2023)Duan, Bian, Cheng, Lyu, Xiong, Xiao, Wang, Duan, Li,
  Huang et~al.}]{duan2023synthesized}
\bibinfo{author}{Duan, C.}, \bibinfo{author}{Bian, X.}, \bibinfo{author}{Cheng,
  K.}, \bibinfo{author}{Lyu, J.}, \bibinfo{author}{Xiong, Y.},
  \bibinfo{author}{Xiao, S.}, \bibinfo{author}{Wang, X.},
  \bibinfo{author}{Duan, Q.}, \bibinfo{author}{Li, C.}, \bibinfo{author}{Huang,
  J.}, et~al., \bibinfo{year}{2023}.
\newblock \bibinfo{title}{Synthesized 7t mprage from 3t mprage using generative
  adversarial network and validation in clinical brain imaging: A feasibility
  study}.
\newblock \bibinfo{journal}{Journal of Magnetic Resonance Imaging} .
\bibitem[{Fischl et~al.(2002)Fischl, Salat, Busa, Albert, Dieterich,
  Haselgrove, Van Der~Kouwe, Killiany, Kennedy, Klaveness
  et~al.}]{fischl2002whole}
\bibinfo{author}{Fischl, B.}, \bibinfo{author}{Salat, D.H.},
  \bibinfo{author}{Busa, E.}, \bibinfo{author}{Albert, M.},
  \bibinfo{author}{Dieterich, M.}, \bibinfo{author}{Haselgrove, C.},
  \bibinfo{author}{Van Der~Kouwe, A.}, \bibinfo{author}{Killiany, R.},
  \bibinfo{author}{Kennedy, D.}, \bibinfo{author}{Klaveness, S.}, et~al.,
  \bibinfo{year}{2002}.
\newblock \bibinfo{title}{Whole brain segmentation: automated labeling of
  neuroanatomical structures in the human brain}.
\newblock \bibinfo{journal}{Neuron} \bibinfo{volume}{33},
  \bibinfo{pages}{341--355}.
\bibitem[{He et~al.(2021)He, Carass, Zuo, Dewey and Prince}]{he2021autoencoder}
\bibinfo{author}{He, Y.}, \bibinfo{author}{Carass, A.}, \bibinfo{author}{Zuo,
  L.}, \bibinfo{author}{Dewey, B.E.}, \bibinfo{author}{Prince, J.L.},
  \bibinfo{year}{2021}.
\newblock \bibinfo{title}{Autoencoder based self-supervised test-time
  adaptation for medical image analysis}.
\newblock \bibinfo{journal}{Medical Image Analysis} , \bibinfo{pages}{102136}.
\bibitem[{Heiss et~al.(2023)Heiss, Weber, Balbach, Schmitt, Rehnitz, Laqmani,
  Sternberg, Ellermann, Nagel, Ladd et~al.}]{heiss2023clinical}
\bibinfo{author}{Heiss, R.}, \bibinfo{author}{Weber, M.A.},
  \bibinfo{author}{Balbach, E.}, \bibinfo{author}{Schmitt, R.},
  \bibinfo{author}{Rehnitz, C.}, \bibinfo{author}{Laqmani, A.},
  \bibinfo{author}{Sternberg, A.}, \bibinfo{author}{Ellermann, J.J.},
  \bibinfo{author}{Nagel, A.M.}, \bibinfo{author}{Ladd, M.E.}, et~al.,
  \bibinfo{year}{2023}.
\newblock \bibinfo{title}{Clinical application of ultrahigh-field-strength
  wrist mri: a multireader 3-t and 7-t comparison study}.
\newblock \bibinfo{journal}{Radiology} \bibinfo{volume}{307},
  \bibinfo{pages}{e220753}.
\bibitem[{Henschel et~al.(2020)Henschel, Conjeti, Estrada, Diers, Fischl and
  Reuter}]{henschel2020fastsurfer}
\bibinfo{author}{Henschel, L.}, \bibinfo{author}{Conjeti, S.},
  \bibinfo{author}{Estrada, S.}, \bibinfo{author}{Diers, K.},
  \bibinfo{author}{Fischl, B.}, \bibinfo{author}{Reuter, M.},
  \bibinfo{year}{2020}.
\newblock \bibinfo{title}{Fastsurfer-a fast and accurate deep learning based
  neuroimaging pipeline}.
\newblock \bibinfo{journal}{NeuroImage} \bibinfo{volume}{219},
  \bibinfo{pages}{117012}.
\bibitem[{Hinton et~al.(2015)Hinton, Vinyals and Dean}]{hinton2015distilling}
\bibinfo{author}{Hinton, G.}, \bibinfo{author}{Vinyals, O.},
  \bibinfo{author}{Dean, J.}, \bibinfo{year}{2015}.
\newblock \bibinfo{title}{Distilling the knowledge in a neural network}.
\newblock \bibinfo{journal}{arXiv preprint arXiv:1503.02531} .
\bibitem[{Hu et~al.(2018)Hu, Shen and Sun}]{hu2018squeeze}
\bibinfo{author}{Hu, J.}, \bibinfo{author}{Shen, L.}, \bibinfo{author}{Sun,
  G.}, \bibinfo{year}{2018}.
\newblock \bibinfo{title}{Squeeze-and-excitation networks}, in:
  \bibinfo{booktitle}{Proceedings of the IEEE/CVF Conference on Computer Vision
  and Pattern Recognition}, pp. \bibinfo{pages}{7132--7141}.
\bibitem[{Huang et~al.(2024)Huang, Yang, Liu, Zhou, Chang, Zhou, Chen, Yu,
  Chen, Chen et~al.}]{huang2024segment}
\bibinfo{author}{Huang, Y.}, \bibinfo{author}{Yang, X.}, \bibinfo{author}{Liu,
  L.}, \bibinfo{author}{Zhou, H.}, \bibinfo{author}{Chang, A.},
  \bibinfo{author}{Zhou, X.}, \bibinfo{author}{Chen, R.}, \bibinfo{author}{Yu,
  J.}, \bibinfo{author}{Chen, J.}, \bibinfo{author}{Chen, C.}, et~al.,
  \bibinfo{year}{2024}.
\newblock \bibinfo{title}{Segment anything model for medical images?}
\newblock \bibinfo{journal}{Medical Image Analysis} \bibinfo{volume}{92},
  \bibinfo{pages}{103061}.
\bibitem[{I{\c{s}}{\i}n et~al.(2016)I{\c{s}}{\i}n, Direko{\u{g}}lu and
  {\c{S}}ah}]{icsin2016review}
\bibinfo{author}{I{\c{s}}{\i}n, A.}, \bibinfo{author}{Direko{\u{g}}lu, C.},
  \bibinfo{author}{{\c{S}}ah, M.}, \bibinfo{year}{2016}.
\newblock \bibinfo{title}{Review of {MRI}-based brain tumor image segmentation
  using deep learning methods}.
\newblock \bibinfo{journal}{Procedia Computer Science} \bibinfo{volume}{102},
  \bibinfo{pages}{317--324}.
\bibitem[{Isola et~al.(2017)Isola, Zhu, Zhou and Efros}]{isola2017image}
\bibinfo{author}{Isola, P.}, \bibinfo{author}{Zhu, J.Y.},
  \bibinfo{author}{Zhou, T.}, \bibinfo{author}{Efros, A.A.},
  \bibinfo{year}{2017}.
\newblock \bibinfo{title}{Image-to-image translation with conditional
  adversarial networks}, in: \bibinfo{booktitle}{Proceedings of the IEEE
  Conference on Computer Vision and Pattern Recognition}, pp.
  \bibinfo{pages}{1125--1134}.
\bibitem[{Jyothi and Singh(2023)}]{jyothi2023deep}
\bibinfo{author}{Jyothi, P.}, \bibinfo{author}{Singh, A.R.},
  \bibinfo{year}{2023}.
\newblock \bibinfo{title}{Deep learning models and traditional automated
  techniques for brain tumor segmentation in mri: a review}.
\newblock \bibinfo{journal}{Artificial intelligence review}
  \bibinfo{volume}{56}, \bibinfo{pages}{2923--2969}.
\bibitem[{Kirillov et~al.(2023)Kirillov, Mintun, Ravi, Mao, Rolland, Gustafson,
  Xiao, Whitehead, Berg, Lo et~al.}]{kirillov2023segment}
\bibinfo{author}{Kirillov, A.}, \bibinfo{author}{Mintun, E.},
  \bibinfo{author}{Ravi, N.}, \bibinfo{author}{Mao, H.},
  \bibinfo{author}{Rolland, C.}, \bibinfo{author}{Gustafson, L.},
  \bibinfo{author}{Xiao, T.}, \bibinfo{author}{Whitehead, S.},
  \bibinfo{author}{Berg, A.C.}, \bibinfo{author}{Lo, W.Y.}, et~al.,
  \bibinfo{year}{2023}.
\newblock \bibinfo{title}{Segment anything}.
\newblock \bibinfo{journal}{arXiv preprint arXiv:2304.02643} .
\bibitem[{Kumar et~al.(2018)Kumar, Nagar, Arora and Gupta}]{kumar2018u}
\bibinfo{author}{Kumar, P.}, \bibinfo{author}{Nagar, P.},
  \bibinfo{author}{Arora, C.}, \bibinfo{author}{Gupta, A.},
  \bibinfo{year}{2018}.
\newblock \bibinfo{title}{U-segnet: fully convolutional neural network based
  automated brain tissue segmentation tool}, in: \bibinfo{booktitle}{2018 25th
  IEEE International Conference on Image Processing (ICIP)},
  \bibinfo{organization}{IEEE}. pp. \bibinfo{pages}{3503--3507}.
\bibitem[{Lee et~al.(2022)Lee, Oh, Shen and Suk}]{lee2022novel}
\bibinfo{author}{Lee, J.}, \bibinfo{author}{Oh, K.}, \bibinfo{author}{Shen,
  D.}, \bibinfo{author}{Suk, H.I.}, \bibinfo{year}{2022}.
\newblock \bibinfo{title}{A novel knowledge keeper network for 7{T}-free but
  7{T}-guided brain tissue segmentation}, in: \bibinfo{booktitle}{International
  Conference on Medical Image Computing and Computer-Assisted Intervention},
  \bibinfo{organization}{Springer}. pp. \bibinfo{pages}{330--339}.
\bibitem[{Li et~al.(2023)Li, Zhang, Zhang, Wang, Ma, Zhang and Wu}]{li2023can}
\bibinfo{author}{Li, Z.}, \bibinfo{author}{Zhang, C.}, \bibinfo{author}{Zhang,
  Y.}, \bibinfo{author}{Wang, X.}, \bibinfo{author}{Ma, X.},
  \bibinfo{author}{Zhang, H.}, \bibinfo{author}{Wu, S.}, \bibinfo{year}{2023}.
\newblock \bibinfo{title}{Can: Context-assisted full attention network for
  brain tissue segmentation}.
\newblock \bibinfo{journal}{Medical Image Analysis} \bibinfo{volume}{85},
  \bibinfo{pages}{102710}.
\bibitem[{Ma and Fu(2020)}]{ma2020position}
\bibinfo{author}{Ma, X.}, \bibinfo{author}{Fu, S.}, \bibinfo{year}{2020}.
\newblock \bibinfo{title}{Position-aware recalibration module: learning from
  feature semantics and feature position}, in:
  \bibinfo{booktitle}{International Joint Conference on Artificial
  Intelligence}, pp. \bibinfo{pages}{797--803}.
\bibitem[{Mao et~al.(2017)Mao, Li, Xie, Lau, Wang and
  Paul~Smolley}]{mao2017least}
\bibinfo{author}{Mao, X.}, \bibinfo{author}{Li, Q.}, \bibinfo{author}{Xie, H.},
  \bibinfo{author}{Lau, R.Y.}, \bibinfo{author}{Wang, Z.},
  \bibinfo{author}{Paul~Smolley, S.}, \bibinfo{year}{2017}.
\newblock \bibinfo{title}{Least squares generative adversarial networks}, in:
  \bibinfo{booktitle}{Proceedings of the IEEE International Conference on
  Computer Vision}, pp. \bibinfo{pages}{2794--2802}.
\bibitem[{Mazurowski et~al.(2023)Mazurowski, Dong, Gu, Yang, Konz and
  Zhang}]{mazurowski2023segment}
\bibinfo{author}{Mazurowski, M.A.}, \bibinfo{author}{Dong, H.},
  \bibinfo{author}{Gu, H.}, \bibinfo{author}{Yang, J.}, \bibinfo{author}{Konz,
  N.}, \bibinfo{author}{Zhang, Y.}, \bibinfo{year}{2023}.
\newblock \bibinfo{title}{Segment anything model for medical image analysis: an
  experimental study}.
\newblock \bibinfo{journal}{Medical Image Analysis} \bibinfo{volume}{89},
  \bibinfo{pages}{102918}.
\bibitem[{Mileti{\'c} et~al.(2022)Mileti{\'c}, Bazin, Isherwood, Keuken,
  Alkemade and Forstmann}]{miletic2022charting}
\bibinfo{author}{Mileti{\'c}, S.}, \bibinfo{author}{Bazin, P.L.},
  \bibinfo{author}{Isherwood, S.J.}, \bibinfo{author}{Keuken, M.C.},
  \bibinfo{author}{Alkemade, A.}, \bibinfo{author}{Forstmann, B.U.},
  \bibinfo{year}{2022}.
\newblock \bibinfo{title}{Charting human subcortical maturation across the
  adult lifespan with in vivo 7{T MRI}}.
\newblock \bibinfo{journal}{NeuroImage} \bibinfo{volume}{249},
  \bibinfo{pages}{118872}.
\bibitem[{Nielsen et~al.(2013)Nielsen, Kinkel, Madigan, Tinelli, Benner and
  Mainero}]{nielsen2013contribution}
\bibinfo{author}{Nielsen, A.S.}, \bibinfo{author}{Kinkel, R.P.},
  \bibinfo{author}{Madigan, N.}, \bibinfo{author}{Tinelli, E.},
  \bibinfo{author}{Benner, T.}, \bibinfo{author}{Mainero, C.},
  \bibinfo{year}{2013}.
\newblock \bibinfo{title}{Contribution of cortical lesion subtypes at 7{T MRI}
  to physical and cognitive performance in {MS}}.
\newblock \bibinfo{journal}{Neurology} \bibinfo{volume}{81},
  \bibinfo{pages}{641--649}.
\bibitem[{Qin et~al.(2021)Qin, Bu, Liu, Shen, Zhou, Gu, Wang, Wu and
  Dai}]{qin2021efficient}
\bibinfo{author}{Qin, D.}, \bibinfo{author}{Bu, J.J.}, \bibinfo{author}{Liu,
  Z.}, \bibinfo{author}{Shen, X.}, \bibinfo{author}{Zhou, S.},
  \bibinfo{author}{Gu, J.J.}, \bibinfo{author}{Wang, Z.H.},
  \bibinfo{author}{Wu, L.}, \bibinfo{author}{Dai, H.F.}, \bibinfo{year}{2021}.
\newblock \bibinfo{title}{Efficient medical image segmentation based on
  knowledge distillation}.
\newblock \bibinfo{journal}{IEEE Transactions on Medical Imaging}
  \bibinfo{volume}{40}, \bibinfo{pages}{3820--3831}.
\bibitem[{Qu et~al.(2020)Qu, Zhang, Wang, Yap and Shen}]{qu2020synthesized}
\bibinfo{author}{Qu, L.}, \bibinfo{author}{Zhang, Y.}, \bibinfo{author}{Wang,
  S.}, \bibinfo{author}{Yap, P.T.}, \bibinfo{author}{Shen, D.},
  \bibinfo{year}{2020}.
\newblock \bibinfo{title}{Synthesized 7{T MRI} from 3{T MRI} via deep learning
  in spatial and wavelet domains}.
\newblock \bibinfo{journal}{Medical Image Analysis} \bibinfo{volume}{62},
  \bibinfo{pages}{101663}.
\bibitem[{Romero et~al.(2014)Romero, Ballas, Kahou, Chassang, Gatta and
  Bengio}]{romero2014fitnets}
\bibinfo{author}{Romero, A.}, \bibinfo{author}{Ballas, N.},
  \bibinfo{author}{Kahou, S.E.}, \bibinfo{author}{Chassang, A.},
  \bibinfo{author}{Gatta, C.}, \bibinfo{author}{Bengio, Y.},
  \bibinfo{year}{2014}.
\newblock \bibinfo{title}{{F}itnets: hints for thin deep nets}.
\newblock \bibinfo{journal}{arXiv preprint arXiv:1412.6550} .
\bibitem[{Ronneberger et~al.(2015)Ronneberger, Fischer and
  Brox}]{ronneberger2015u}
\bibinfo{author}{Ronneberger, O.}, \bibinfo{author}{Fischer, P.},
  \bibinfo{author}{Brox, T.}, \bibinfo{year}{2015}.
\newblock \bibinfo{title}{U-{N}et: convolutional networks for biomedical image
  segmentation}, in: \bibinfo{booktitle}{International Conference on Medical
  Image Computing and Computer-Assisted Intervention},
  \bibinfo{organization}{Springer}. pp. \bibinfo{pages}{234--241}.
\bibitem[{Roy et~al.(2019)Roy, Conjeti, Navab, Wachinger, Initiative
  et~al.}]{roy2019quicknat}
\bibinfo{author}{Roy, A.G.}, \bibinfo{author}{Conjeti, S.},
  \bibinfo{author}{Navab, N.}, \bibinfo{author}{Wachinger, C.},
  \bibinfo{author}{Initiative, A.D.N.}, et~al., \bibinfo{year}{2019}.
\newblock \bibinfo{title}{Quicknat: a fully convolutional network for quick and
  accurate segmentation of neuroanatomy}.
\newblock \bibinfo{journal}{NeuroImage} \bibinfo{volume}{186},
  \bibinfo{pages}{713--727}.
\bibitem[{Shaffer et~al.(2022)Shaffer, Kwok, Naik, Anderson, Lam, Wszalek,
  Arnold and Hassaneen}]{shaffer2022ultra}
\bibinfo{author}{Shaffer, A.}, \bibinfo{author}{Kwok, S.S.},
  \bibinfo{author}{Naik, A.}, \bibinfo{author}{Anderson, A.T.},
  \bibinfo{author}{Lam, F.}, \bibinfo{author}{Wszalek, T.},
  \bibinfo{author}{Arnold, P.M.}, \bibinfo{author}{Hassaneen, W.},
  \bibinfo{year}{2022}.
\newblock \bibinfo{title}{Ultra-high-field {MRI} in the diagnosis and
  management of gliomas: a systematic review}.
\newblock \bibinfo{journal}{Frontiers in Neurology} \bibinfo{volume}{13},
  \bibinfo{pages}{857825--857825}.
\bibitem[{Svanera et~al.(2021)Svanera, Benini, Bontempi and
  Muckli}]{svanera2021cerebrum}
\bibinfo{author}{Svanera, M.}, \bibinfo{author}{Benini, S.},
  \bibinfo{author}{Bontempi, D.}, \bibinfo{author}{Muckli, L.},
  \bibinfo{year}{2021}.
\newblock \bibinfo{title}{{CEREBRUM}-7{T}: fast and fully volumetric brain
  segmentation of 7 {T}esla {MR} volumes}.
\newblock \bibinfo{journal}{Human Brain Mapping} \bibinfo{volume}{42},
  \bibinfo{pages}{5563--5580}.
\bibitem[{Van~Reeth et~al.(2012)Van~Reeth, Tham, Tan and Poh}]{van2012super}
\bibinfo{author}{Van~Reeth, E.}, \bibinfo{author}{Tham, I.W.},
  \bibinfo{author}{Tan, C.H.}, \bibinfo{author}{Poh, C.L.},
  \bibinfo{year}{2012}.
\newblock \bibinfo{title}{Super-resolution in magnetic resonance imaging: a
  review}.
\newblock \bibinfo{journal}{Concepts in Magnetic Resonance Part A}
  \bibinfo{volume}{40}, \bibinfo{pages}{306--325}.
\bibitem[{Wang et~al.(2020)Wang, Li, Wang, Hu and Yang}]{wang2020collaborative}
\bibinfo{author}{Wang, H.}, \bibinfo{author}{Li, Y.}, \bibinfo{author}{Wang,
  Y.}, \bibinfo{author}{Hu, H.}, \bibinfo{author}{Yang, M.H.},
  \bibinfo{year}{2020}.
\newblock \bibinfo{title}{Collaborative distillation for ultra-resolution
  universal style transfer}, in: \bibinfo{booktitle}{Proceedings of the
  IEEE/CVF Conference on Computer Vision and Pattern Recognition}, pp.
  \bibinfo{pages}{1860--1869}.
\bibitem[{Wang et~al.(2019)Wang, Xie and Zeng}]{wang2019rp}
\bibinfo{author}{Wang, L.}, \bibinfo{author}{Xie, C.}, \bibinfo{author}{Zeng,
  N.}, \bibinfo{year}{2019}.
\newblock \bibinfo{title}{R{P}-{N}et: a 3{D} convolutional neural network for
  brain segmentation from magnetic resonance imaging}.
\newblock \bibinfo{journal}{IEEE Access} \bibinfo{volume}{7},
  \bibinfo{pages}{39670--39679}.
\bibitem[{Wang et~al.(2021)Wang, Li, Singh, Lu and
  Vasconcelos}]{wang2021imagine}
\bibinfo{author}{Wang, P.}, \bibinfo{author}{Li, Y.}, \bibinfo{author}{Singh,
  K.K.}, \bibinfo{author}{Lu, J.}, \bibinfo{author}{Vasconcelos, N.},
  \bibinfo{year}{2021}.
\newblock \bibinfo{title}{I{MAGINE}: Image synthesis by image-guided model
  inversion}, in: \bibinfo{booktitle}{Proceedings of the IEEE/CVF Conference on
  Computer Vision and Pattern Recognition}, pp. \bibinfo{pages}{3681--3690}.
\bibitem[{Wei et~al.(2021)Wei, Wu, Wang, Bui, Qu, Yap, Xia, Li and
  Shen}]{wei2021cascaded}
\bibinfo{author}{Wei, J.}, \bibinfo{author}{Wu, Z.}, \bibinfo{author}{Wang,
  L.}, \bibinfo{author}{Bui, T.D.}, \bibinfo{author}{Qu, L.},
  \bibinfo{author}{Yap, P.T.}, \bibinfo{author}{Xia, Y.}, \bibinfo{author}{Li,
  G.}, \bibinfo{author}{Shen, D.}, \bibinfo{year}{2021}.
\newblock \bibinfo{title}{A cascaded nested network for 3{T} brain {MR} image
  segmentation guided by 7{T} labeling}.
\newblock \bibinfo{journal}{Pattern Recognition} , \bibinfo{pages}{108420}.
\bibitem[{Yu et~al.(2023)Yu, Yang, Zhou, Cai, Gao, Lee, Li, Bao, Xu, Lasko
  et~al.}]{yu2023unest}
\bibinfo{author}{Yu, X.}, \bibinfo{author}{Yang, Q.}, \bibinfo{author}{Zhou,
  Y.}, \bibinfo{author}{Cai, L.Y.}, \bibinfo{author}{Gao, R.},
  \bibinfo{author}{Lee, H.H.}, \bibinfo{author}{Li, T.}, \bibinfo{author}{Bao,
  S.}, \bibinfo{author}{Xu, Z.}, \bibinfo{author}{Lasko, T.A.}, et~al.,
  \bibinfo{year}{2023}.
\newblock \bibinfo{title}{Unest: Local spatial representation learning with
  hierarchical transformer for efficient medical segmentation}.
\newblock \bibinfo{journal}{Medical Image Analysis} \bibinfo{volume}{90},
  \bibinfo{pages}{102939}.
\bibitem[{Zhang et~al.(2001)Zhang, Brady and Smith}]{zhang2001segmentation}
\bibinfo{author}{Zhang, Y.}, \bibinfo{author}{Brady, M.},
  \bibinfo{author}{Smith, S.}, \bibinfo{year}{2001}.
\newblock \bibinfo{title}{Segmentation of brain {MR} images through a hidden
  {M}arkov random field model and the expectation-maximization algorithm}.
\newblock \bibinfo{journal}{IEEE Transactions on Medical Imaging}
  \bibinfo{volume}{20}, \bibinfo{pages}{45--57}.
\bibitem[{Zhang et~al.(2018)Zhang, Cheng, Xiang, Yap and Shen}]{zhang2018dual}
\bibinfo{author}{Zhang, Y.}, \bibinfo{author}{Cheng, J.Z.},
  \bibinfo{author}{Xiang, L.}, \bibinfo{author}{Yap, P.T.},
  \bibinfo{author}{Shen, D.}, \bibinfo{year}{2018}.
\newblock \bibinfo{title}{Dual-domain cascaded regression for synthesizing 7{T}
  from 3{T MRI}}, in: \bibinfo{booktitle}{International Conference on Medical
  Image Computing and Computer-Assisted Intervention},
  \bibinfo{organization}{Springer}. pp. \bibinfo{pages}{410--417}.

\end{thebibliography}





\end{document}